\def\year{2021}\relax
\documentclass[letterpaper]{article} 
\usepackage{arxiv}  
\usepackage{times}  
\usepackage{helvet} 
\usepackage{courier}  
\usepackage[hyphens]{url}  
\usepackage{graphicx} 
\usepackage{natbib}  
\usepackage{caption} 
\newcommand{\eg}{\textit{e.g.}}
\newcommand{\ie}{\textit{i.e.}}
\usepackage{color}
\usepackage[caption = false]{subfig}
\usepackage{bm}
\usepackage{amsmath,amssymb}

\title{Understanding Deformable Alignment in Video Super-Resolution}
\author {
	Kelvin C.K. Chan\textsuperscript{\rm 1}\footnote{Both authors contributed equally to this work.}\,\,
	Xintao Wang\textsuperscript{\rm 2}\footnotemark[1]\footnote{This work was done during his study in CUHK.}\,\,
	Ke Yu\textsuperscript{\rm 3}\,\,
	Chao Dong\textsuperscript{\rm 4}\,\,
	Chen Change Loy\textsuperscript{\rm 1}\footnote{Corresponding author.}\\
}
\affiliations {
	\textsuperscript{\rm 1}Nanyang Technological University, Singapore\quad
	\textsuperscript{\rm 2}Applied Research Center, Tencent PCG\\
    \textsuperscript{\rm 3}CUHK -- SenseTime Joint Lab, The Chinese University of Hong Kong\\
    \textsuperscript{\rm 4}SIAT -- SenseTime Joint Lab, Shenzhen Institutes of Advanced Technology, Chinese Academy of Sciences\\
	\textsuperscript{\rm 1}\{chan0899, ccloy\}@ntu.edu.sg\,\,
	\textsuperscript{\rm 2}xintao.wang@outlook.com\,\,
	\textsuperscript{\rm 3}yk017@ie.cuhk.edu.hk\,\,
	\textsuperscript{\rm 4}chao.dong@siat.ac.cn
}

\begin{document}

\maketitle

\begin{abstract}
    Deformable convolution, originally proposed for the adaptation to geometric variations of objects, has recently shown compelling performance in aligning multiple frames and is increasingly adopted for video super-resolution. Despite its remarkable performance, its underlying mechanism for alignment remains unclear.
    In this study, we carefully investigate the relation between deformable alignment and the classic flow-based alignment. We show that deformable convolution can be decomposed into a combination of spatial warping and convolution. This decomposition reveals the commonality of deformable alignment and flow-based alignment in formulation, but with a key difference in their offset diversity. We further demonstrate through experiments that the increased diversity in deformable alignment yields better-aligned features, and hence significantly improves the quality of video super-resolution output.
    Based on our observations, we propose an offset-fidelity loss that guides the offset learning with optical flow.
    Experiments show that our loss successfully avoids the overflow of offsets and alleviates the instability problem of deformable alignment.
    Aside from the contributions to deformable alignment, our formulation inspires a more flexible approach to introduce offset diversity to flow-based alignment, improving its performance.
\end{abstract}

\section{Introduction}
Video super-resolution (SR) aims at recovering high-resolution consecutive frames from their low-resolution counterparts.
The key challenge of video SR lies in the effective use of complementary details from adjacent frames, which can be misaligned due to camera and object motions.
To establish inter-frame correspondence, early methods~\cite{caballero2017real,liu2017robust,sajjadi2018frame,tao2017detail,xue2019video} employ optical flow for explicit frame alignment. They warp neighboring frames to the reference one, and pass these images to Convolutional Neural Networks (CNNs) for super-resolution. Recent studies~\cite{tian2018tdan,wang2019deformable,wang2019edvr} perform the alignment operation implicitly via deformable convolution and show superior performance. For instance, the winner of NTIRE 2019 video restoration challenges~\cite{nah2019ntire_deblur,nah2019ntire_sr}, EDVR~\cite{wang2019edvr}, significantly outperforms previous methods with coarse-to-fine deformable convolutions.

\begin{figure*}[!t]
	\begin{center}
		\includegraphics[width=0.9\textwidth]{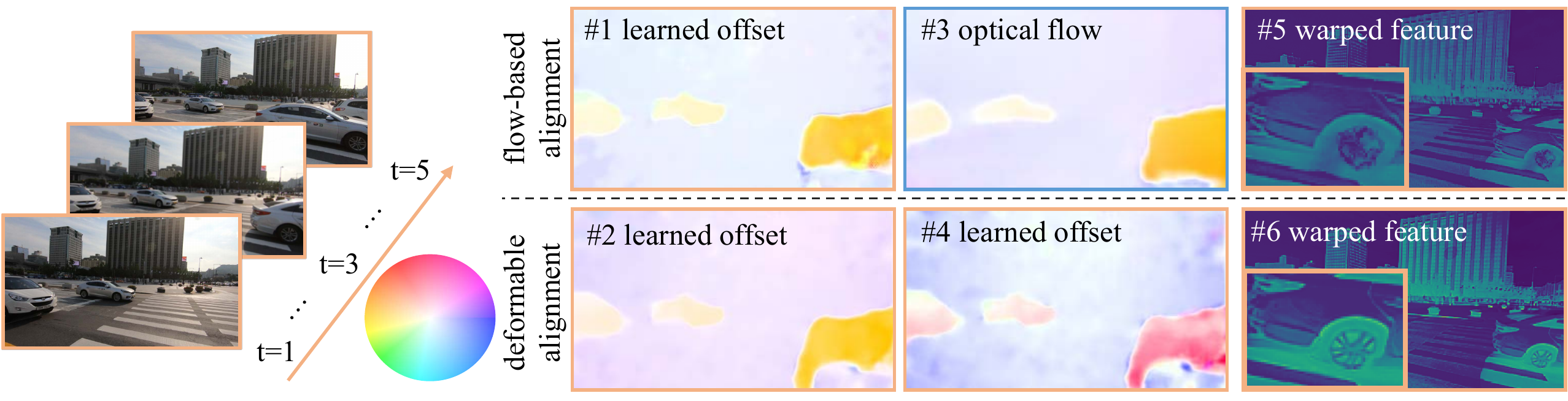}
		\caption{The learned offsets in both flow-based alignment (\#1) and deformable alignment (\#2, \#4) have similar patterns as optical flow obtained using a deep learning-based optical flow estimator~\cite{sun2018pwc} (\#3). The offset diversity allows deformale alignment to learn complementary offsets (\#4), which effectively alleviate the occulusion problem and reduce warping errors. As a result, the warped feature after deformable alignment (\# 6) contains more details than that with flow-based alignment (\# 5) (see details of the car wheel).}
		\label{fig:teaser}
	\end{center}
\end{figure*}

These two kinds of methods are generally regarded as orthogonal approaches and are developed independently. It is of great interest to know (1) \textit{the relationship between explicit and implicit alignments}, and (2) \textit{the source of improvement brought by implicit modeling}. As there are no related works, we bridge the gap by exploring the intrinsic connections of two representative methods --  flow-based alignment (explicit alignment with optical flow) and deformable alignment (implicit alignment with deformable convolution). Studying their relation not only helps us understand the working mechanism of deformable alignment, but also inspires a more general design of video SR approaches.

Deformable convolution~\cite{dai2017deformable,zhu2019deformable} (DCN) is originally designed for spatial adaption in object detection. The key idea is to displace the sampling locations of standard convolution by some learned offsets. When DCN is applied in temporal alignment, the displaced kernels on neighboring frames will be used to align intermediate features. On the face of it, this procedure is different from flow-based methods, which align adjacent frames by flow-warping. To reveal their relationship, we show that deformable alignment can be formulated as a combination of feature-level flow-warping and convolution. This intuitive decomposition indicates that these two kinds of alignment intrinsically share the same formulation but differ in their \textit{offset diversity}. Specifically, flow-based alignment only learns one offset at each feature location, while deformable alignment introduces multiple offsets, the number of which is in proportion to the kernel size of DCN.

Under this relation, we systematically investigate the effects of offset diversity and gain two interesting insights.
First, the learned offsets in deformable alignment have similar patterns as optical flow, suggesting that deformable and flow-based alignments are strongly correlated in both concepts and behaviors.
Second, diverse offsets achieve better restoration performance than a single offset. As different offsets are complementary to each other, they can effectively alleviate the occlusion problem and reduce warping errors caused by large motions. Figure~\ref{fig:teaser} depicts the comparisons of these two methods in their learned offsets and feature patterns.

With a more profound understanding of their relationship, we decided to use the widely-adopted optical flow technique to benefit the training of deformable convolution. It is known that the training of deformable alignment is unstable and the overflow of offsets could severely degrade the performance~\cite{wang2019edvr}.
We propose an offset-fidelity loss that adopts optical flows to guide the offset learning of DCN while preserving offset diversity. Our experiments show that the proposed strategy successfully stabilizes the training process of deformable alignment.

Apart from the contributions to deformable alignment, our decomposition of DCN is also beneficial to flow-based alignment approaches. Specifically, in our formulation, the number of offsets is not necessarily equal to the square of kernel size. Compared to deformable convolution, our formulation provides a more flexible means for increasing offset diversity in flow-based alignment approaches.

Our \textbf{contributions} are summarized as follows:
(1) While deformable alignment has been shown a compelling alternative to the conventional flow-based alignment for motion compensation, its link with flow-based alignment is only superficially discussed in the literature. This paper is the first study that establishes the relationship between the two important concepts formally.
(2) We systematically investigate the benefits of offset diversity. We show that offset diversity is the key factor for improving both the alignment accuracy and SR performance.
(3) Based on our studies, we propose an offset-fidelity loss in deformable alignment to stabilize training while preserving offset diversity. An improvement of up to 1.7~dB is observed with our loss.
(4) Our formulation inspires a more flexible approach to increase offset diversity in flow-based alignment methods.

\section{Related Work}
Different from single image SR~\cite{dai2019secondorder,dong2014learning,haris2018deep,he2019ode,ledig2017photo,lim2017enhanced,liu2020residual,mei2020image,wang2018esrgan,wang2018recovering,zhang2018image,zhang2020deep}, an additional challenge of video SR~\cite{dai2015dictionary,huang2015bidirectional,liu2014bayesian,takeda2009super,yi2019progressive,li2020mucan,isobe2020video1,isobe2020video} is to align multiple frames for the construction of accurate correspondences.
Based on whether optical flow is explicitly estimated, existing motion compensation approaches in video SR can be mainly divided into two branches -- \textit{explicit} methods and \textit{implicit} methods.

Most existing methods adopt an explicit motion compensation approach. Earlier works of this approach~\cite{kappeler2016video,liao2015video} first use a fixed and external optical flow estimator to estimate the flow fields between the reference and its neighboring frames, and then learn a mapping from the flow-warped inputs to the high-resolution output. Such two-stage methods are time-consuming and tend to fail when the flow estimation is not accurate. Several follow-up studies~\cite{caballero2017real,liu2017robust,sajjadi2018frame,tao2017detail,xue2019video} incorporate the flow-estimation component into the SR pipeline. For instance, TOFlow~\cite{xue2019video} points out that the optimal flow is task-specific in video enhancement including video SR, and thus a trainable motion estimation component is more effective than a fixed one. Nevertheless, all these methods explicitly perform flow estimation and warping in the \textit{image domain}, which may introduce artifacts around image structures~\cite{tian2018tdan}.

Several recent methods perform motion compensation implicitly and show superior performance. For instance, DUF~\cite{jo2018deep} learns an upsampling filter for each pixel location, and a few other methods~\cite{tian2018tdan,wang2019deformable,wang2019edvr} incorporate deformable convolution into motion compensation. Deformable convolution~\cite{dai2017deformable} is capable of predicting additional offsets that offer spatial flexibilities to a convolution kernel. This differs from a standard convolution, which is restricted to a regular neighborhood.
TDAN~\cite{tian2018tdan} applies deformable convolutions~\cite{dai2017deformable} for temporal alignment in video SR. Following the structure design in flow-estimation methods~\cite{dosovitskiy2015flownet,ranjan2017optical,sun2018pwc}, EDVR~\cite{wang2019edvr} adopts deformable alignment in a pyramid and cascading architecture, and achieves the state-of-the-art performance in video SR.

Although deformable alignment and the classic flow-based alignment look unconnected at first glance, they are indeed highly related.
In this study, we delve deep into the connections between them. Based on our analyses, we propose an offset-fidelity loss to stabilize the training and improve the performance of deformable alignment.

\section{Unifying Deformable and  Flow-Based Alignments}
\label{sec:decomp}
\begin{figure}[!t]
	\begin{center}
		\subfloat[DCN]{\includegraphics[height=0.15\textwidth]{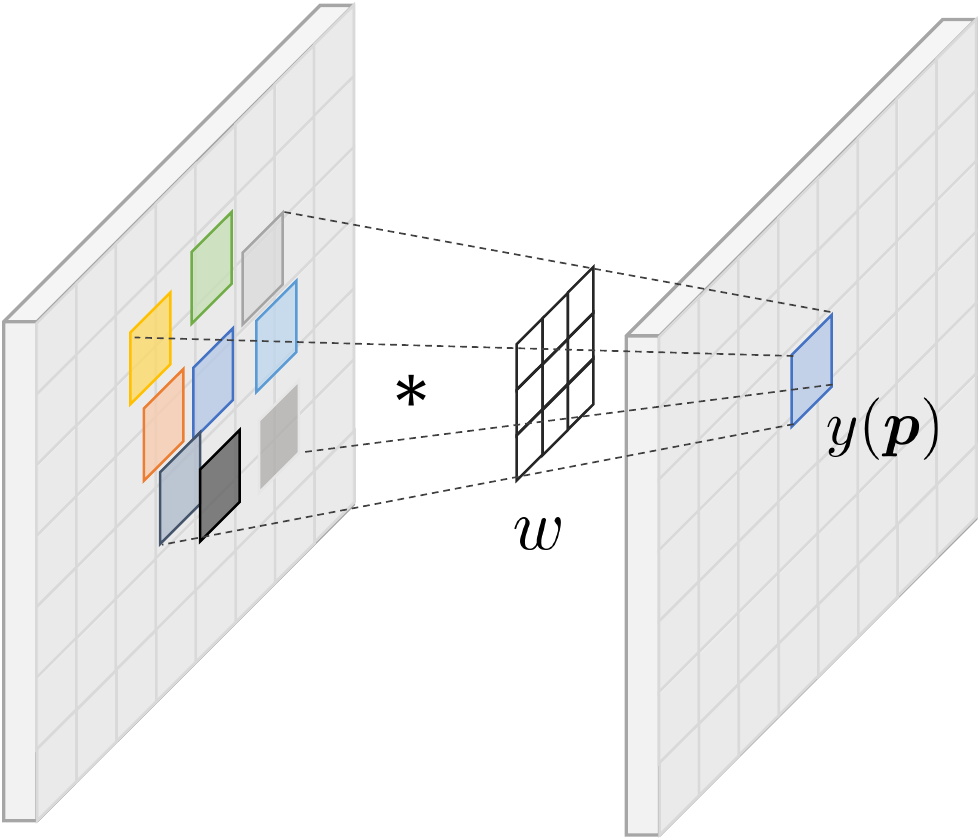}}\quad
		\subfloat[DCN Decomposition]{\includegraphics[height=0.15\textwidth]{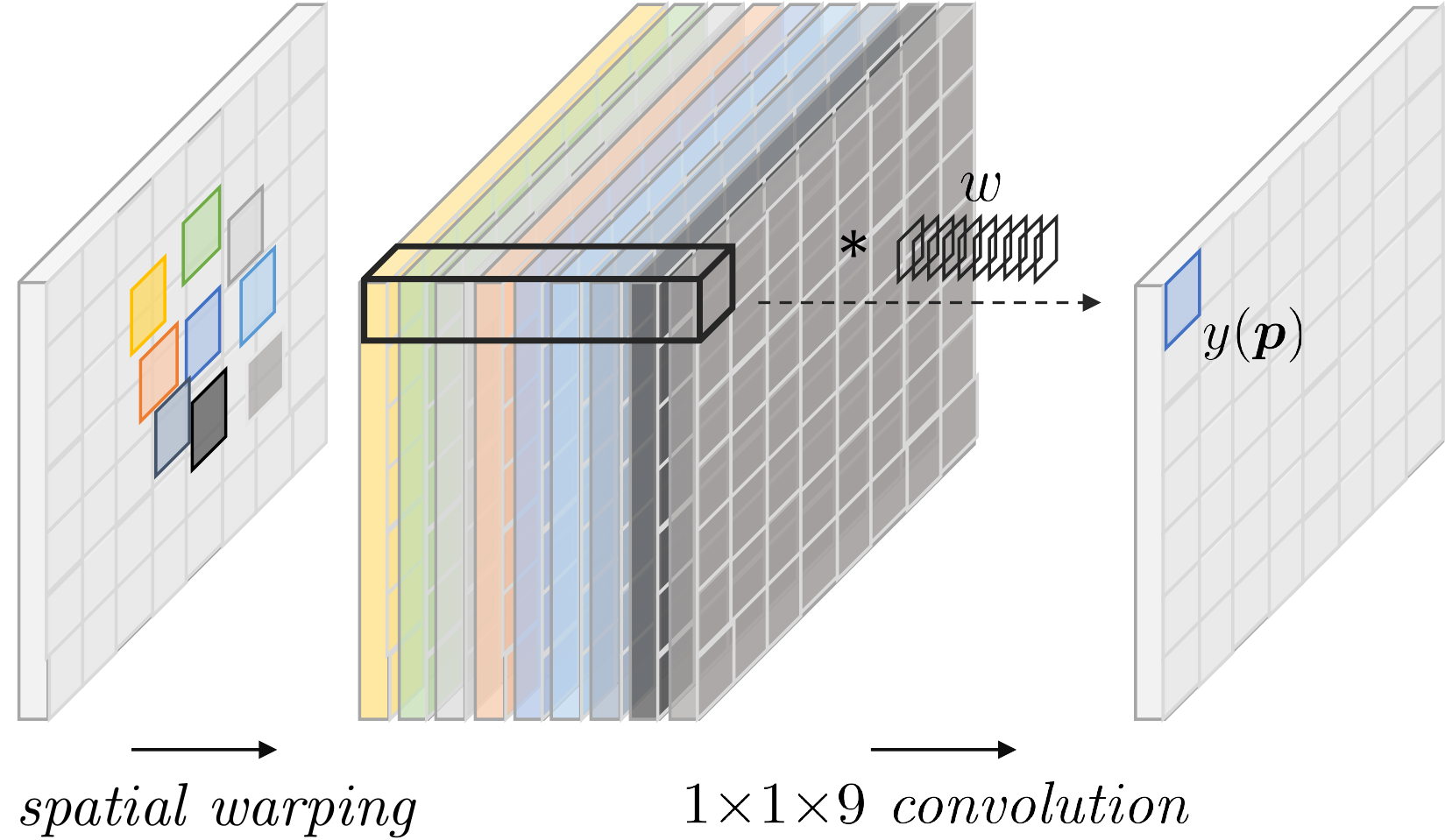}}
		\caption{Deformable convolution with a $3{\times3}$ kernel can be decomposed into nine spatial warpings and one 3D convolution. Kernel weights are represented as $w$.}
		\label{fig:deformable}
	\end{center}
\end{figure}

\subsection{Deformable Convolution Revisited}
We start with a brief review of deformable convolution (DCN)~\cite{dai2017deformable}, which was originally proposed to accommodate geometric variations of objects in the tasks of object detection~\cite{bertasius2018object} and image segmentation~\cite{dai2017deformable}.
Let $\bm{p}_k$ be the $k$-th sampling offset in a standard convolution with kernel size $n {\times} n$. For example, when $n{=}3$, we have $\bm{p}_k\, {\in}\,\{(-1,-1), (-1,0), \cdots, (1,1)\}$. We denote the $k$-th additional learned offset at location $\bm{p} + \bm{p}_k$ by $\Delta\bm{p}_k$. A deformable convolution can be formulated as:
\begin{equation}
\label{eq:dcn}
	y(\bm{p}) = \sum_{k=1}^{n^2} w(\bm{p}_k) \cdot x(\bm{p}+\bm{p}_k+\Delta\bm{p}_k),
\end{equation}
where $x$ and $y$ represent the input and output features, respectively. The kernel weights are denoted by $w$.
As illustrated in Fig.~\ref{fig:deformable}(a), unlike standard convolution, a deformable convolution has more flexible sampling locations.
In practice, one can divide the $C$ channel features into $G$ groups of features with $C/G$ channels, and $n^2 {\times} G$ offsets are learned for each spatial location. In DCNv2~\cite{zhu2019deformable} a modulation mask is introduced to further strengthen the capability in manipulating spatial support regions. A detailed analysis of the modulation mask is included in Sec.~\ref{appx:masks}.

\subsection{Deformable Alignment}
\begin{figure}[!t]
		\begin{center}
		\includegraphics[width=0.35\textwidth]{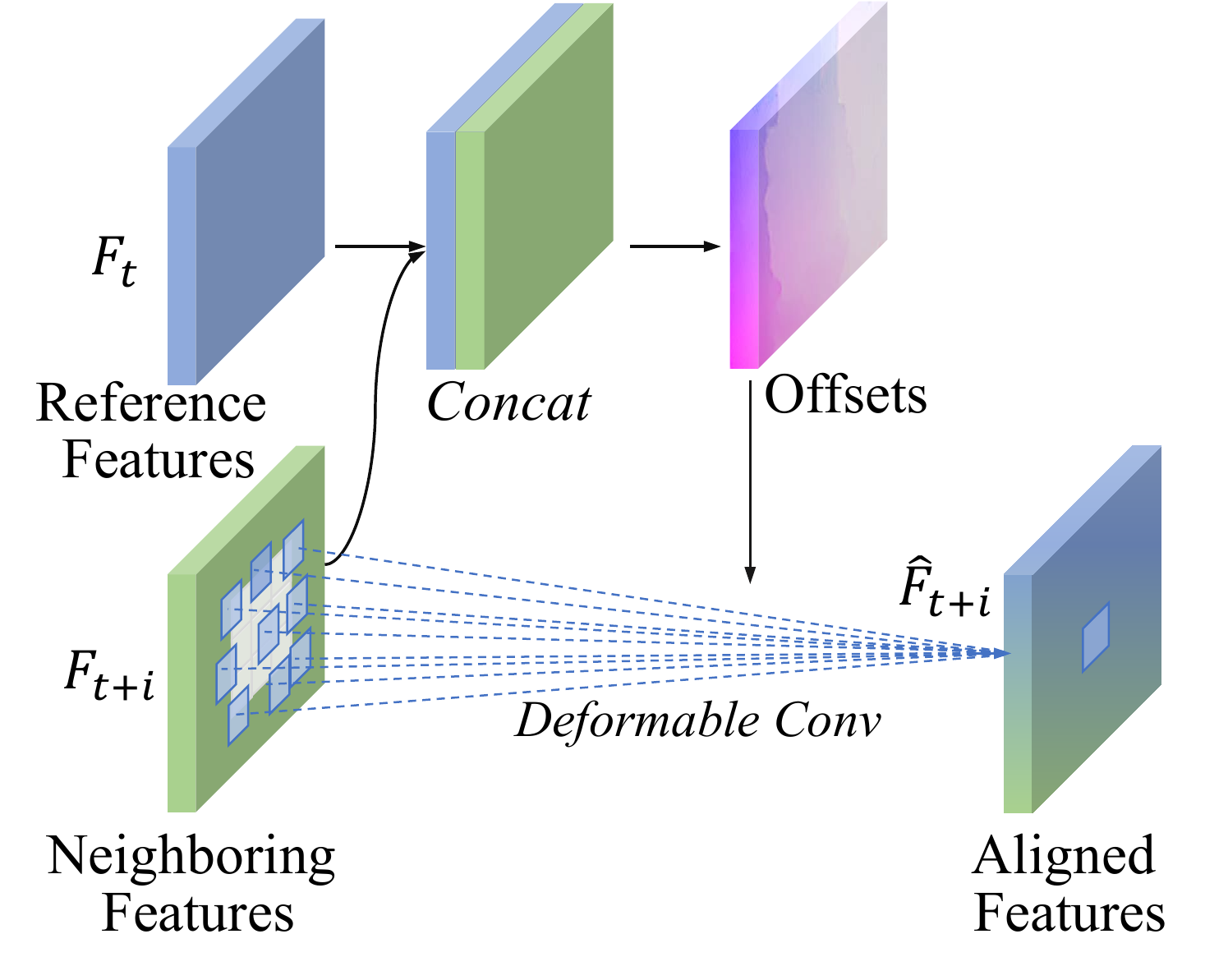}
		\caption{Deformable alignment applies deformable convolution to align neighboring features to the reference feature. The offsets are predicted by a few convolutions with both the reference and neighboring features as the inputs. The reference feature is used only to predict the offsets, and is not directly involved in the convolution.}
		\label{fig:deformable_alignment}
		\end{center}
\end{figure}

In video SR, it is crucial to establish correspondences between consecutive frames for detail extraction and fusion.
Recent studies~\cite{tian2018tdan,wang2019deformable,wang2019edvr} go beyond the traditional way of flow-warping and apply deformable convolution for feature alignment, as shown in Fig.~\ref{fig:deformable_alignment}.

Let $F_t$ and $F_{t+i}$ be the intermediate features of the reference and neighboring frames, respectively. In deformable alignment, a deformable convolution is used to align $F_{t+i}$ to $F_t$. Mathematically, we have:
\begin{equation}
	\hat{F}_{t+i}(\bm{p}) = \sum_{k=1}^{n^2} w(\bm{p}_k) \cdot F_{t+i}(\bm{p}+\bm{p}_k+\Delta\bm{p}_k),
\end{equation}
where $\hat{F}_{t+i}$ represents the aligned feature. The offsets $\Delta\bm{p}_k$ are predicted by a few convolutions with both $F_t$ and $F_{t+i}$ as the inputs.
The reference feature is used only to predict the offsets, and is not directly involved in the convolution.

\subsection{Relation between Deformable Alignment and Flow-Based Alignment}
\label{subsec:proof}
There is an intuitive yet less obvious connection between deformable and flow-based alignments. The connection is rarely discussed in previous works. Instead of treating them as orthogonal approaches, in this paper we unify these two important concepts. Next, we discuss the connection between deformable alignment and flow-based alignment by showing that DCN can be decomposed into spatial warping and standard convolution.

Let $x$ be the input feature, and $\bm{p}_k + \Delta \bm{p}_k$ ($k=1,\cdots,n^2$) be the $k$-th offset for location $\bm{p}$. Next, denote the feature warped by the $k$-th offset by $x_k(\bm{p}) = x(\bm{p} + \bm{p}_k + \Delta \bm{p}_k)$. From Eqn.~(\ref{eq:dcn}), we have:
	\begin{equation}
	\label{eq:w3c}
		y(\bm{p}) = \sum_{k = 1}^{n^2} w(\bm{p}_k) \cdot x_k(\bm{p}),
	\end{equation}
which is equivalent to a $1 {\times} 1 {\times} n^2$ standard 3D convolution.
Hence, we see that a deformable convolution with kernel size $n {\times} n$ is equivalent to $n^2$ individual spatial warpings followed by a standard 3D convolution with kernel size $1 {\times} 1 {\times} n^2$. The illustration is shown in Fig.~\ref{fig:deformable}(b).
\vspace{2cm}

\noindent\textbf{Remarks:} \begin{enumerate}
	\item By replacing $n^2$ with $N{\in}\,\mathbb{N}$ in Eqn.~(\ref{eq:w3c}) , this decomposition \textit{generalizes} DCN by removing the constraint that the number of offsets within each group must be equal to $n^2$. Therefore, in the remaining sections, we denote the number of offsets per group by $N$.
	\item By stacking the $N$ warped features in the channel dimension, the $1 {\times} 1 {\times} N$ 3D convolution can be implemented as a $1 {\times} 1$ 2D convolution. In other words, DCN is equivalent to $N$ separate spatial warpings followed by a $1{\times1}$ 2D convolution.
\end{enumerate}

{\noindent}From Eqn.~(\ref{eq:w3c}), we see that the special case of $n{=}1$ is equivalent to a spatial warping followed by a $1 {\times} 1$ convolution. In the context of motion compensation, this special case corresponds to a flow-based alignment. In other words, deformable and flow-based alignments share the same formulation but with a difference in the \textit{offset diversity}.\\

\noindent\textbf{Discussion.}
The aforementioned analysis leads to a few interesting explorations:
\begin{enumerate}
	\item \textit{Where does deformable alignment gain the extra performance in comparison to flow-based alignment?} The analysis points to offset diversity, and we verify this hypothesis in our experiments in Sec.~\ref{sec:analyses}.
	\item \textit{Is higher offset diversity always better?} We demonstrate in Sec.~\ref{subsec:compare} that although the output quality increases with offset diversity in general, a performance plateau is observed when the number of offsets gets larger. Hence, indefinitely increasing the number of offsets could lower the efficiency of the model without significant performance gain. In practice, one should balance the performance and computational efficiency by choosing a suitable number of offsets.
	\item \textit{Can we increase the offset diversity of flow-based alignment?} Unlike deformable alignment, where the number of offsets must be equal to the square of kernel size, our formulation generalizes deformable alignment with an arbitrary number of offsets. As a result, it provides a more flexible approach to introduce offset diversity to flow-based alignment. We show in the experiments that increasing offset diversity helps a flow-based network to achieve better SR performance.
\end{enumerate}

\subsection{Offset-fidelity Loss}
In this section, motivated by the decomposition shown in Sec.~\ref{subsec:proof}, we demonstrate how optical flow can benefit deformable alignment through the newly proposed offset fidelity loss.

Due to its unclear offset interpretability, deformable alignment is usually trained from scratch with random initializations.
With increased network capacities, the training of deformable alignment becomes unstable, and the overflow of offsets severely degrades the model performance\footnote{The instability of EDVR is observed in~\cite{wang2019edvr} and also in our experiments.}.
In contrast, in flow-based alignment, various training strategies are developed to improve alignment accuracy and speed of convergence, such as the adoption of flow network structure~\cite{haris2019recurrent,xue2019video}, flow guidance loss~\cite{liu2017robust}, and flow pre-training~\cite{caballero2017real,tao2017detail,xue2019video}.

Given the relation between spatial warping and deformable convolution discussed in Sec.~\ref{subsec:proof}, we proposed to use optical flow to guide the training of offsets.
Specifically, we propose an offset-fidelity loss to constrain the offsets so that they do not deviate much from the optical flow. Furthermore, to facilitate the learning of optimal and diverse offsets for video SR, Heaviside step function is incorporated. More specifically, we augment the data-fitting loss as follows:
\begin{equation}
\hat{L} = L + \lambda\sum_{n=1}^N L_n,
\end{equation}
where $L$ is the data-fitting loss (\eg, Charbonnier loss in~\cite{wang2019edvr}) and
\begin{equation}
L_n = \sum_{i}\sum_j H\left(|x_{n, ij} - y_{ij}| - t\right)\cdot|x_{n, ij} - y_{ij}|,
\end{equation}
where $i, j$ denote the spatial indices and $H(\cdot)$ represents the Heaviside step function. Here $\lambda$ and $t$ are hyper-parameters controling the diversity of the offsets.
As shown in Sec.~\ref{subsec:offset-fidelity}, our loss is able to stabilize the training and avoid the offset overflow in large models.

\section{Analysis}
\label{sec:analyses}
We conduct experiments to reveal the connections and differences between deformable and flow-based alignments in Video SR.
Unless specified, EDVR-M\footnote{EDVR-M is a moderate version of EDVR provided by the official implementation~\cite{wang2019edvr}.} is adopted for analyses as it maintains a good balance between training efficiency and performance.
Moreover, to decouple the complex relation among different components in deformable alignment, we use a non-modulated DCN in Sec.~\ref{subsec:flowbased} and Sec.~\ref{subsec:compare}. The details of each experiment are provided in Sec.~\ref{appx:settings}.

\subsection{Deformable Alignment vs. Optical Flow }
\label{subsec:flowbased}
\begin{figure}[!t]
	\begin{center}
		\includegraphics[width=0.45\textwidth]{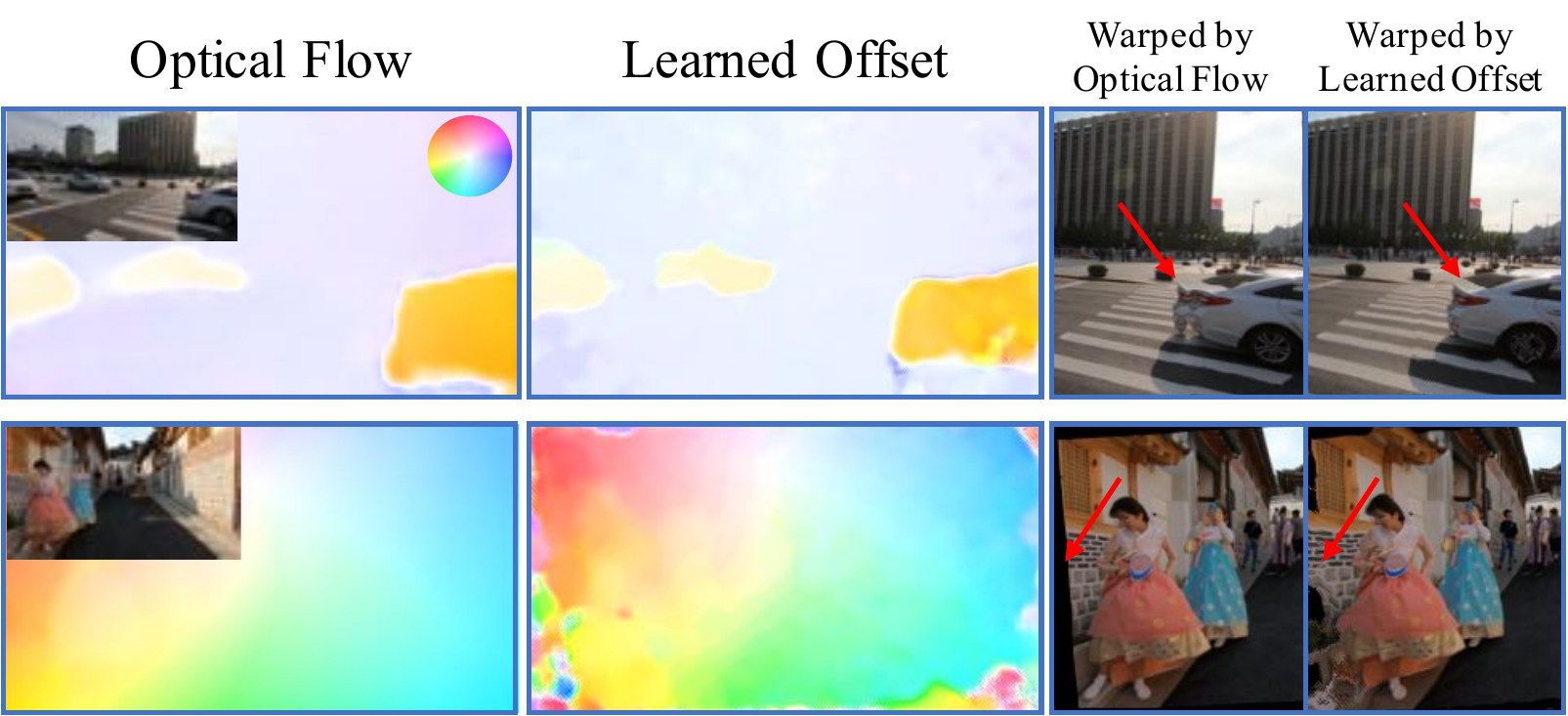}
		\caption{While the learned offsets are highly similar to optical flows, their disparity is non-negligible because optical flow may not be optimal for video SR. Dark borders and ghosting regions appear in the images warped by optical flow. In contrast, those warped by learned offsets are clearer.}
		\label{fig:flow_comp}
	\end{center}
\end{figure}

By setting $G{=}N{=}1$ (\textit{i.e.}, group=$1$ and the number of offsets per group=$1$), the offset learned by deformable alignment resembles that captured by optical flow in a flow-based alignment approach.
Specifically, when there is only one offset to learn, the model automatically learns to align features based on the motions between frames. As shown in Fig.~\ref{fig:flow_comp}, the learned offsets are highly similar to the optical flows estimated by PWC-Net~\cite{sun2018pwc}.

Despite their high similarity, the disparity between learned offsets and optical flows is non-negligible due to the fundamental difference of the task nature~\cite{xue2019video}. Specifically, while PWC-Net is trained to describe the motions between frames, our baseline is trained for video SR, in which optical flow may not be the optimal representation of frame correspondences. From Fig.~\ref{fig:flow_comp}, we see that the image warped by the learned offsets clearly preserves more scene contents. In contrast, a dark region and a ghosting region are seen in the images warped by optical flow. Note that the offsets are learned for warping the features, and the warped images in Fig.~\ref{fig:flow_comp} are solely for illustrative purposes.

\begin{figure}[!t]
\begin{center}
	\hspace{-0.5cm}\includegraphics[width=0.5\textwidth]{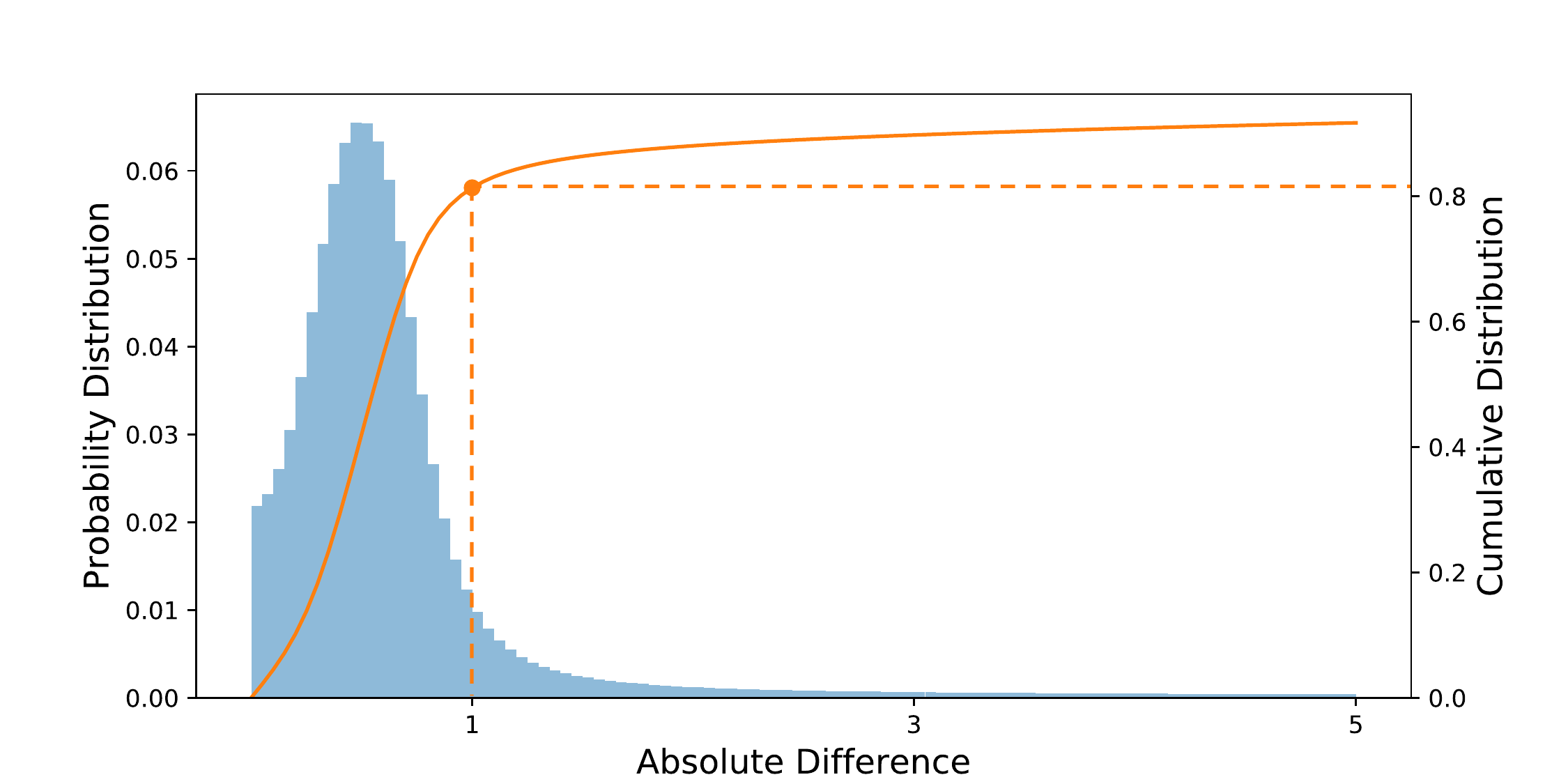}
	\caption{Over 80\% of the estimations have a difference to optical flow smaller than 1 pixel (orange dot). This demonstrates that in the case of $G{=}N{=}1$, deformable alignment is indeed equivalent to flow-based alignment.}
	\label{fig:flow_distri}
\end{center}
\end{figure}

We quantitatively study the correlation between the offsets and optical flows, by computing their pixelwise difference.
As shown in Fig.~\ref{fig:flow_distri}, over 80\% of the estimations have a difference smaller than one pixel from the optical flow.
This demonstrates that in the case of $G{=}N{=}1$, deformable alignment is indeed highly equivalent to flow-based alignment. In the following analyses, we will adopt this model as our approximation to flow-based alignment baseline.

\subsubsection{Feature Warping.}
The aforementioned flow-based alignment baseline performs feature warping. This differs from a majority of flow-based methods that learn flows for image warping~\cite{liu2017robust,xue2019video}. In those methods, the flows contain fractional values and hence interpolation is required during warping. This inevitably introduces information loss, particularly high-frequency details. Consequently, the blurry aligned images yield suboptimal SR results.
Recent deformable alignment methods~\cite{tian2018tdan,wang2019deformable,wang2019edvr} attempt to perform alignment at the feature level and achieve remarkable results.
We inspect the contribution of feature-level warping by replacing the feature alignment module in our flow-based baseline by an image alignment module. Surprisingly, despite the closeness of the architecture, image alignment leads to a drop of 0.84~dB. This indicates that feature-level warping is beneficial to flow-based alignment. More comparisons are shown in Sec.~\ref{appx:warping}.

\subsection{Offset Diversity}
\label{subsec:compare}
\subsubsection{Decomposition Equivalence.}
In this section, we use our decomposition in Sec.~\ref{subsec:proof} in place of DCN since it provides a more flexible choice of the number of offsets. To verify their equivalence, we train two instantiations -- original DCN and our decomposition. As shown in Table~\ref{tab:equiv}, our experiments show that the two instantiations achieve similar performance, corroborating our hypothesis.\\

 \begin{table}[!t]
 	\small
 	\caption{PSNR of two instantiations of EDVR-M on REDS4. The similar performance verifies our claim that DCN can be decomposed into spatial warping and convolution. $G$ and $k$ represent the number of groups and the kernel size used in DCN, respectively.}
 	\label{tab:equiv}
 	\begin{center}
 		\tabcolsep=0.1cm
 		\scalebox{1}{
 			\begin{tabular}{r|c|c|c|c}
 				&$G{=}1,k{=}1$&$G{=}1,k{=}3$&$G{=}8,k{=}1$&$G{=}8,k{=}3$\\
 				\hline
 				DCN&29.979&30.199&30.183&30.264\\
 				Our Decomp.&29.992&30.240&30.179&30.231\\\hline
 				Difference&+0.013&+0.041&-0.004&-0.033
 		\end{tabular}}
 	\end{center}
 \end{table}

\noindent\textbf{Learned Offsets.}
Given that the primary difference between flow-based alignment and deformable alignment is the number of offsets $N$, it is natural to question \textit{the roles and characteristics of the additional offsets in deformable alignment
(\ie\,$N{>}1$).}
To answer this, we fix $G{=}1$ and compare the performance in the cases of $N{=}1$ and $N{=}15$.

\begin{figure}[!t]
	\begin{center}
		\hspace{-0.5cm}
		\includegraphics[width=0.45\textwidth]{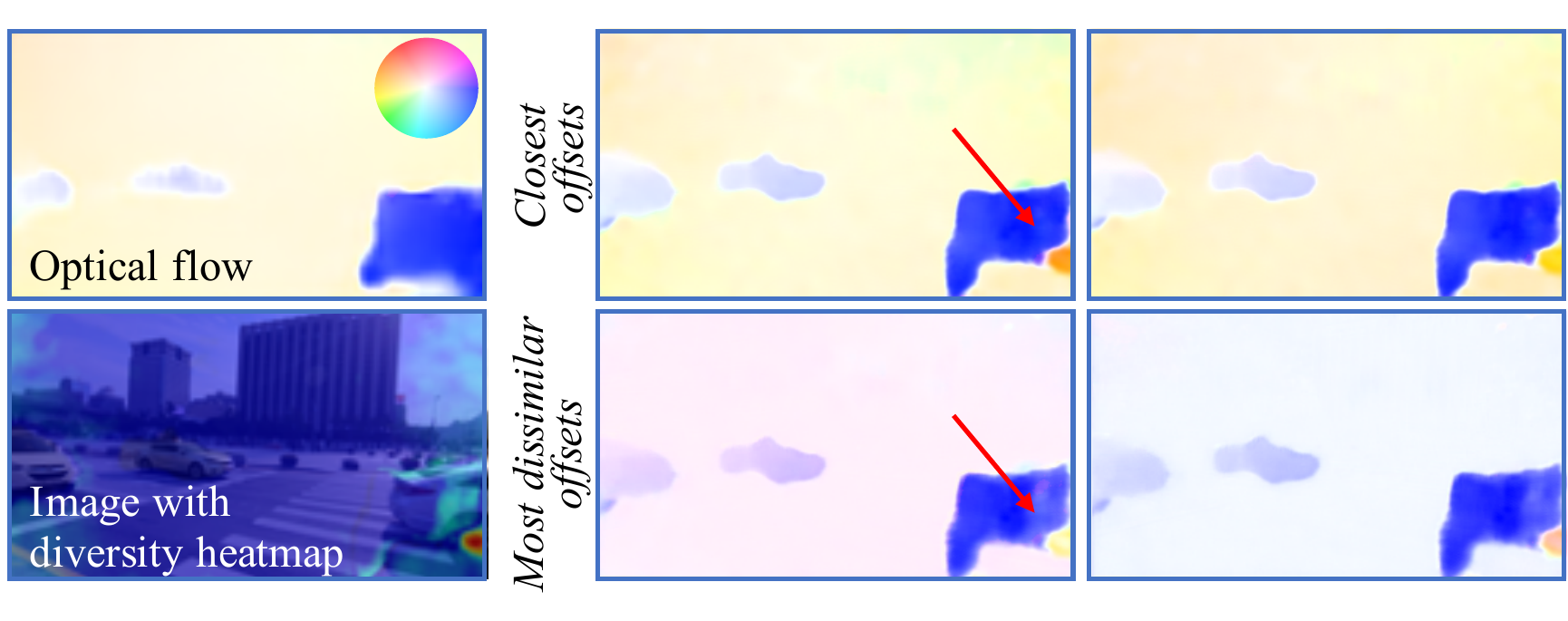}
		\caption{Offsets sorted according to their $l_1$-distance to optical flow. While the closest offsets are highly similar to optical flow with slight differences (top row), the most dissimilar offsets have different estimated directions (bottom row). Moreover, the offsets tend to have a larger diversity in regions that optical flows do not work well for alignment (see the diversity heatmap and regions marked by red arrows). (\textbf{Zoom in for best view})}
		\label{fig:G1N15_visualize}
	\end{center}
\end{figure}

We sort the $15$ offsets according to their $l_1$-distance to the optical flow, and an example is shown in Fig.~\ref{fig:G1N15_visualize}.
On the one hand, there exist offsets that closely resemble the optical flow.
On the other hand, there are offsets that have different estimated directions compared to the optical flows; although these offsets are also able to separate the motions of different objects, as the optical flow does, their directions do not correspond to the actual camera and object motions.

We further visualize the diversity of the offsets, which is measured by the pixelwise standard deviation of the offsets. We observe that the offsets tend to have a larger diversity in regions that optical flows do not work well for alignment. For instance, as shown in the heatmap of Fig.~\ref{fig:G1N15_visualize}, the standard deviation tends to be larger in image boundaries, where unseen regions are common.
Although a diverse set of offsets with different estimated directions is obtained, they are all analogous to optical flow in terms of their overall shape. This suggests that the motion between frames is still an important clue in deformable alignment, as in flow-based alignment. More qualitative results are shown in Sec.~\ref{appx:offsets}

\begin{table}[!t]
	\small
	\caption{PSNR on Vimeo-90K-T~\cite{xue2019video} using two additional models. For both models, the performance increases with increased number of offsets.}
	\label{tab:others_models}
	\begin{center}
		\tabcolsep=0.2cm
		\scalebox{1}{
			\begin{tabular}{c|c|c}
				\multicolumn{3}{c}{\textbf{TDAN}\vspace{0.1cm}}\\
				$N{=}1$&$N{=}9$&$N{=}25$\\\hline
				33.313&33.483&\textbf{33.540}
			\end{tabular}\hspace{0.7cm}
			\begin{tabular}{c|c|c}
				\multicolumn{3}{c}{\textbf{Flow-based}\vspace{0.1cm}}\\
				$N{=}1$&~$N{=}2$&$N{=}3$\\\hline
				32.835&32.973&\textbf{33.017}
			\end{tabular}}
		\end{center}
\end{table}

\begin{figure}[!t]
	\begin{center}
		\includegraphics[width=0.47\textwidth]{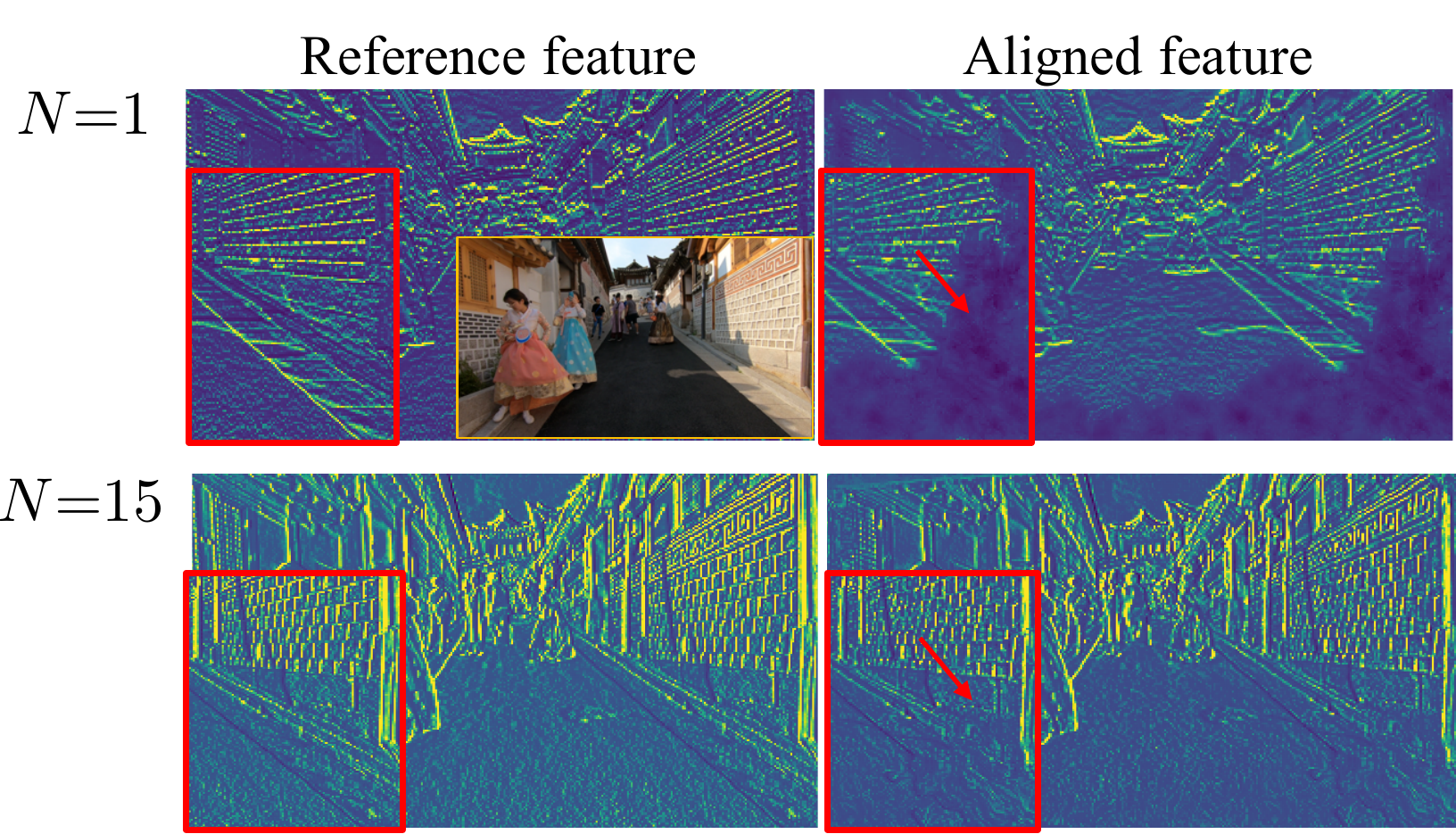}
		\caption{Aligned features with $N{=}1$ and $N{=}15$. With a single offset, the model lacks the ability to handle occlusion and inaccurate motion estimation (red arrows). With increased number of offsets, the features are better aligned, and more details are better preserved. (\textbf{Zoom in for best view})}
		\label{fig:G1N15_aligned}
	\end{center}
\end{figure}

\noindent\textbf{Contributions of Diversity.}
We are also interested in \textit{whether the diverse flow-like offsets are beneficial to video SR}.
This motivates us to inspect the aligned features and the corresponding performance.
With a single offset, the aligned features suffer from the warping error induced by unseen regions and inaccurate motion estimation. The inaccurately aligned features inevitably hinder the aggregation of information and therefore harm the subsequent restoration.
In contrast, with multiple offsets, the independently warped features are reciprocal and provide better aligned features during fusion, hence alleviating the inaccurate alignment by a single offset.
Two examples of the aligned features are visualized in Fig.~\ref{fig:G1N15_aligned}.
It is observed that with a single offset, the aligned features are less coherent. For instance, in the image boundaries, which correspond to the regions that do not exist in the neighboring frame, the feature warped by a single offset contains a large area of dark regions.
Contrarily, with 15 offsets, the complementary warped features provide additional information for fusion, resulting in features that are more coherent and preserve more details.

\noindent\textbf{Increasing Offset Diversity.}
We then examine the performance gain by gradually increasing the number of offsets and attempt to examine \textit{if more offsets will always lead to a better performance}.

\begin{figure*}[!t]
	\begin{center}
		\includegraphics[width=0.95\textwidth]{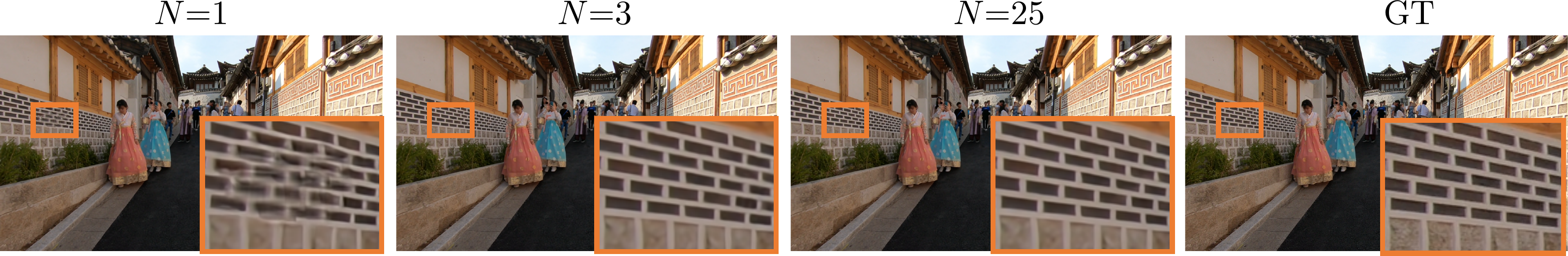}
		\caption{Qualitative comparisons. While the quality improves markedly from $N{=}1$ to $N{=}3$, further improvement at $N{=}25$ is relatively small.}
		\label{fig:qualitative_comp}
	\end{center}
\end{figure*}

\begin{figure}[!t]
	\begin{center}
		\hspace{-0.5cm}\includegraphics[width=0.5\textwidth]{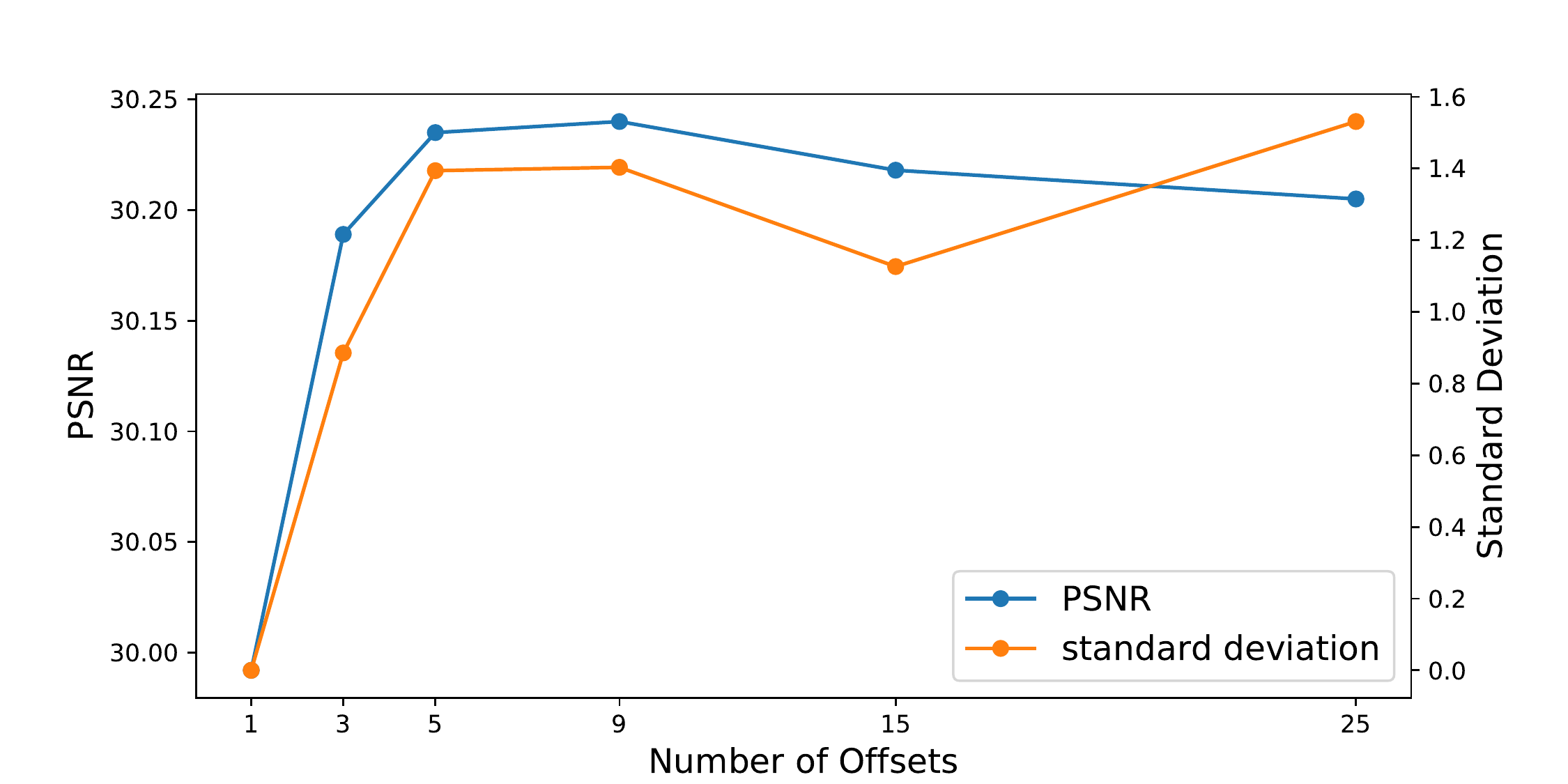}
		\caption{PSNR and offset diversity measured by standard deviation. The performance of the model positively correlates with the offset diversity. When $N$ increases from 1 to 5, both the PSNR and standard deviation of the offsets increase rapidly. When $N$ further increases, both the PSNR and the standard deviation of the offsets saturate.}
    	\label{fig:G1_PSNR}
	\end{center}
\end{figure}

The qualitative and quantitative comparison with different $N$ are shown in Fig.~\ref{fig:qualitative_comp} and Fig.~\ref{fig:G1_PSNR}, respectively. In particular, as the number of offsets increases from 1 to 5, the PSNR increases rapidly. When $N$ further increases, the PSNR saturates at about 30.23~dB.
This result indicates that the performance reaches a plateau when the number of offsets gets larger. As a result, simply increasing the number of offsets could lower the computational efficiency without significant performance gain.
It is worth noting that it is infeasible to balance the performance and computational
efficiency in deformable alignment, since the number of offsets must be equal to the square of kernel size. Our formulation, contrarily, generalizes deformable alignment with an arbitrary number of offsets, thus providing a more flexible approach to introducing offset diversity.

We also inspect the correlation between offset diversity and PSNR performance. We measure the offset diversity by the pixelwise standard deviation of all offsets.
As shown in Fig.~\ref{fig:G1_PSNR}, the performance of the model positively correlates with the offset diversity (Pearson Correlation Coefficient=0.9418 based on these six data points). This result implies that offset diversity actually contributes to the performance gain.

To further support our conclusion, we additionally test the improvement brought by offset diversity using TDAN~\cite{tian2018tdan} and a flow-based network\footnote{We use an architecture similar to TOFlow~\cite{xue2019video} except that spatial warping is done on LR space instead of HR space for computational efficiency. We increase the number of offsets by using multiple SPyNet~\cite{ranjan2017optical}.}. As shown in Table~\ref{tab:others_models}, the PSNR of the two models improves by up to 0.23~dB.
Besides, an improvement of 0.18~dB is observed in the flow-based network, suggesting that the offset diversity not only improves feature alignment, but is also constructive in image alignment.

Besides increasing the number of offsets $N$, offset diversity can also be achieved by increasing the number of group $G$. Interestingly, the above conclusions for $N$ are also applicable to $G$. A more detailed analysis is included in the Sec.~\ref{appx:groups}.

\subsection{Offset-fidelity Loss}
\label{subsec:offset-fidelity}
\begin{figure}[!t]
	\begin{center}
		\hspace{-0.5cm}
		\subfloat[REDS]{\includegraphics[width=0.5\textwidth]{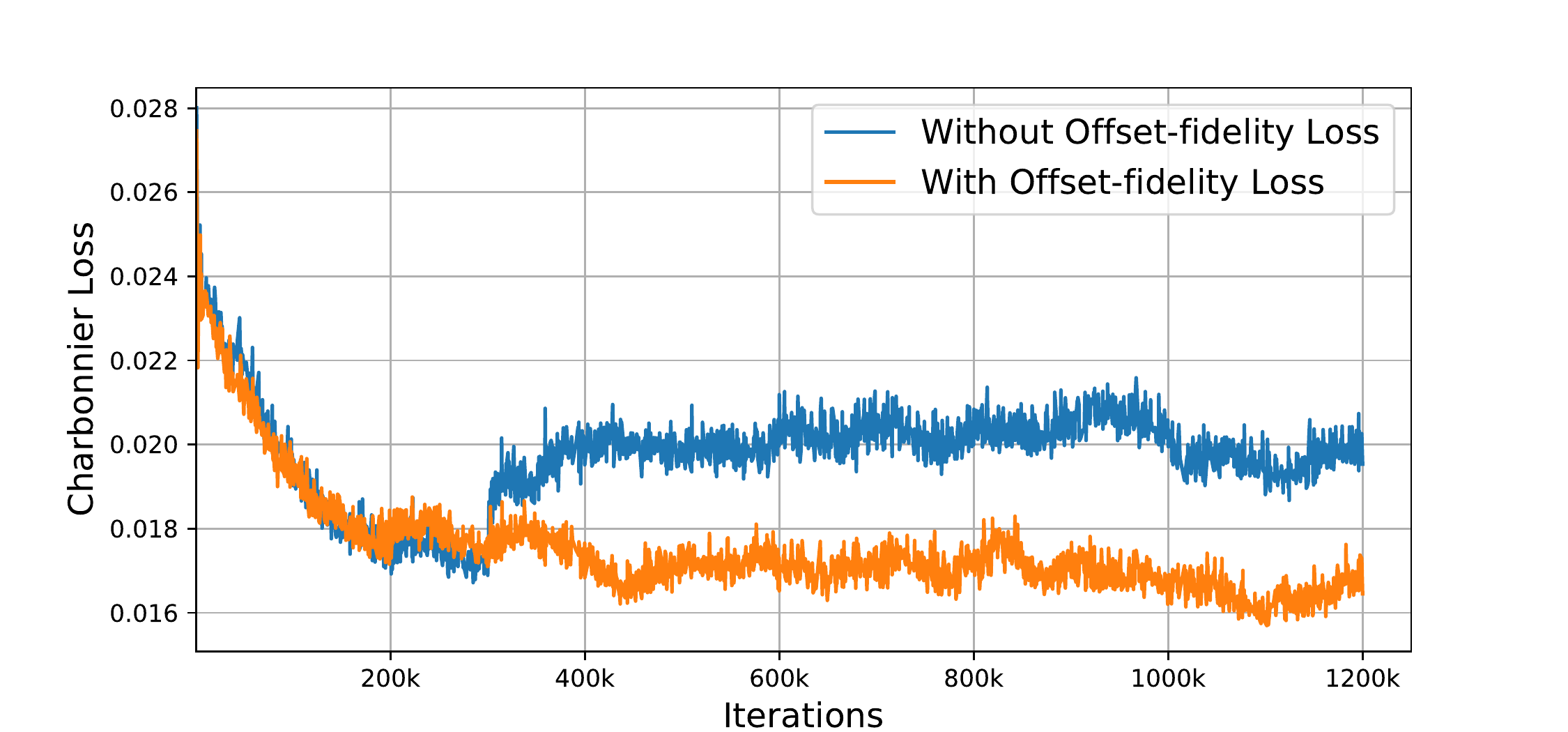}}\\
		\hspace{-0.5cm}
		\subfloat[Vimeo90K]{\includegraphics[width=0.5\textwidth]{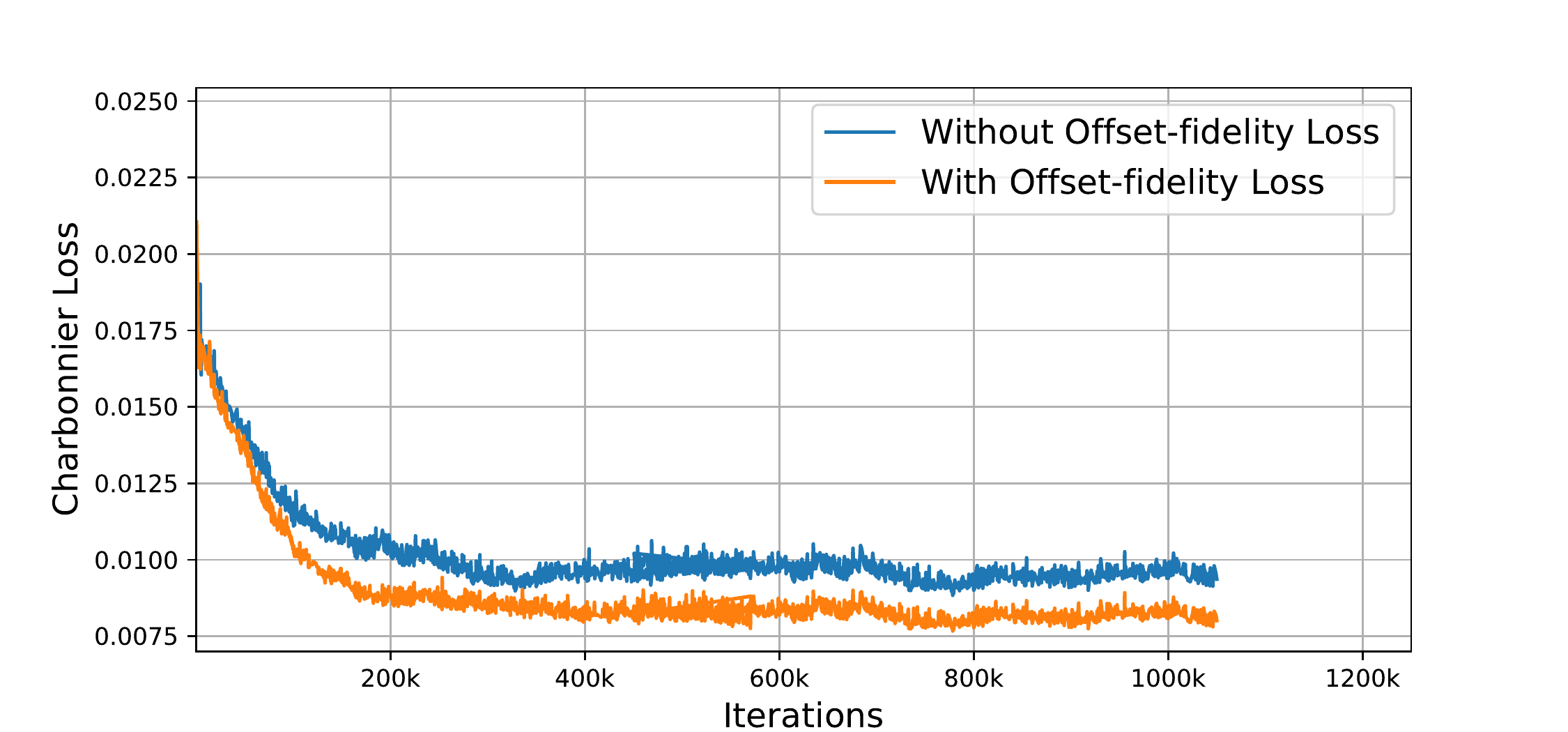}}
		\caption{The training curves of the models. (a) When training on REDS, the model trained without the offset-fidelity loss becomes unstable at about 300K, where the loss increases and is consistently greater than that trained with our loss thereafter. (b) When training on Vimeo-90K, the model trained with offset-fidelity loss is able to reach a lower training loss.}
		\label{fig:loss_curve}
	\end{center}
\end{figure}

We train EDVR-L with the official training scheme. As the network capacity increases, the training of deformable alignment becomes unstable.
Without the offset-fidelity loss, the overflow of offsets produces a zero feature map after deformable alignment. As a result, EDVR essentially becomes a single image SR model.
On the contrary, our loss penalizes the offsets when they deviate from the optical flow, resulting in much more interpretable offsets and a better performance.
As shown in Fig.~\ref{fig:loss_curve}, EDVR is able to converge with a lower training loss with our offset-fidelity loss. Note that in Fig.~\ref{fig:loss_curve}(a), the training loss increases at about 300K, which is the time when offsets overflow. In Table~\ref{tab:loss_output} we see that our loss introduces an additional improvement of up to 1.73~dB. The qualitative results are provided in Sec.~\ref{appx:loss}.

\begin{table}[!t]
	\caption{Quantitative comparison (PSNR) on REDS4 and Vimeo-90K-T for 4${\times}$ video super-resolution. Results are evaluated on RGB channels.}
	\label{tab:loss_output}
	\begin{center}
		\tabcolsep=0.25cm
		\scalebox{1}{
			\begin{tabular}{r|c|c}
				&REDS4&Vimeo-90K-T\\\hline
				without offset-fidelity loss&28.753&33.632\\
				with offset-fidelity loss&30.480&35.223\\\hline
				Difference&+1.727&+1.591
			\end{tabular}}
	\end{center}
\end{table}
\section{Conclusion}
The success of deformable alignment in video super-resolution has aroused great attention. In this study, we uncover the intrinsic connection in both concepts and behaviors between deformable alignment and flow-based alignment.
For flow-based alignment, our work relaxes the constraint of deformable convolution on the number of offsets. It allows a more flexible way to increase the offset diversity in flow-based alignment approaches, improving the output quality.
As for deformable alignment, our investigation empowers us to understand its underlying mechanism, potentially inspiring new alignment approaches. Motivated by our analysis, we propose an offset-fidelity loss to mitigate the stability problem during training.

\appendix
\section{Appendix}
\subsection{Experimental Settings}
\label{appx:settings}
We consider two widely-used datasets: REDS dataset~\cite{nah2019ntire} and Vimeo-90K~\cite{xue2019video}.
REDS is a dataset proposed in the NTIRE19 video restoration challenge~\cite{nah2019ntire}. It contains a total of 240 training, 30 validation, 30 test clips, each having 100 frames with resolution $1280{\times}720$. Following~\cite{wang2019edvr}, we use the REDS4 dataset (Clips 000, 011, 015, 020) as our test set, and the remaining 266 training and validation clips for training.
Vimeo-90K contains 91,701 7-frame sequences with fixed resolution $448{\times}250$. Its testing set, Vimeo-90K-T, is used for testing.

The training scheme mostly follows the EDVR~\cite{wang2019edvr}.
More specifically, we adopt Adam optimizer~\cite{kingma2014adam} and Cosine Annealing with restarts scheme~\cite{loshchilov2016sgdr}. The total number of iterations is $6 {\times} 10^{5}$; the initial learning rate is $4 {\times} 10^{-4}$. The batch size is 16 and the LR patch size is $64{\times}64$. We perform random flipping and rotations for data augmentation.

The training is divided into two stages. First, a variant without the TSA module is pretrained. Then, the pretrained weights are used to initialize the full model (with TSA module). Note that the pretrained weights are frozen at the first 50k iterations. Charbonnier penalty function~\cite{charbonnier1994two} is used as the loss function. We implement the models with PyTorch and train them using four NVIDIA Tesla V100 GPUs.

\subsection{Learned Offsets in Deformable Alignment}
\label{appx:offsets}
\begin{figure*}[h]
	\begin{center}
		\includegraphics[width=0.86\textwidth]{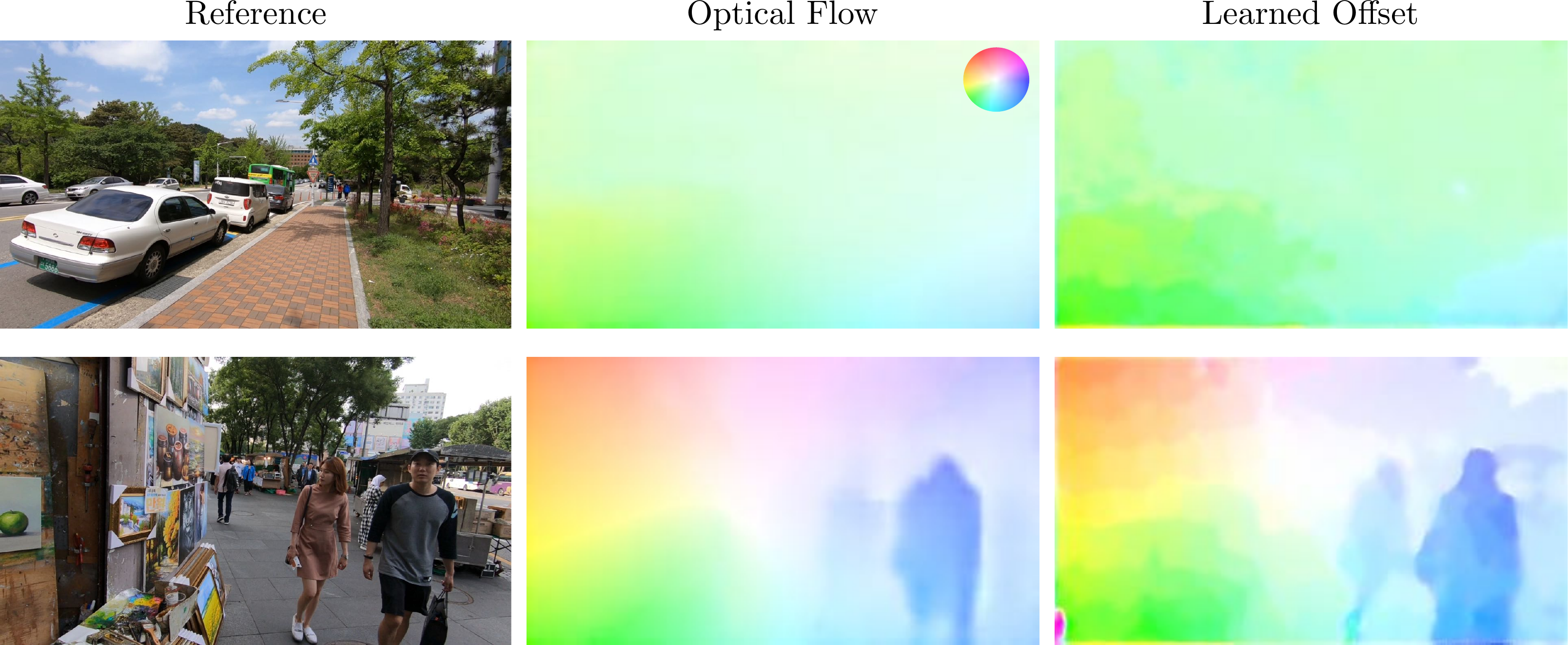}
		\caption{The learned offsets in the case of $G{=}N{=}1$. The learned offsets are highly similar to the optical flow computed by PWCNet~\cite{sun2018pwc}. Moreover, the offsets trained specifically for video SR has non-negligible differences to optical flow.}
		\label{appxfig:G1N1_flow}
	\end{center}
\end{figure*}

As discussed in Sec.~\ref{subsec:flowbased}, when $G{=}N{=}1$, deformable alignment and flow-based alignment are similar to each other, except that flow-based alignment is performed on image level. Here we provide another example about their learned offsets in Fig.~\ref{appxfig:G1N1_flow}. We see that the learned offsets are highly similar to the optical flow by PWC-Net~\cite{sun2018pwc}. In addition, we found that when trained for video SR, the offsets and the optical flow have a non-negligible difference. This observation is explained in~\cite{xue2019video} that the optimal flow should be task-specific. For example, in Fig.~\ref{appxfig:G1N1_flow} (bottom), the learned offsets for the human is similar to the optical flow, but the boundaries in the learned offsets is sharper.

\begin{figure*}[h]
	\begin{center}
		\subfloat{\includegraphics[width=0.88\textwidth]{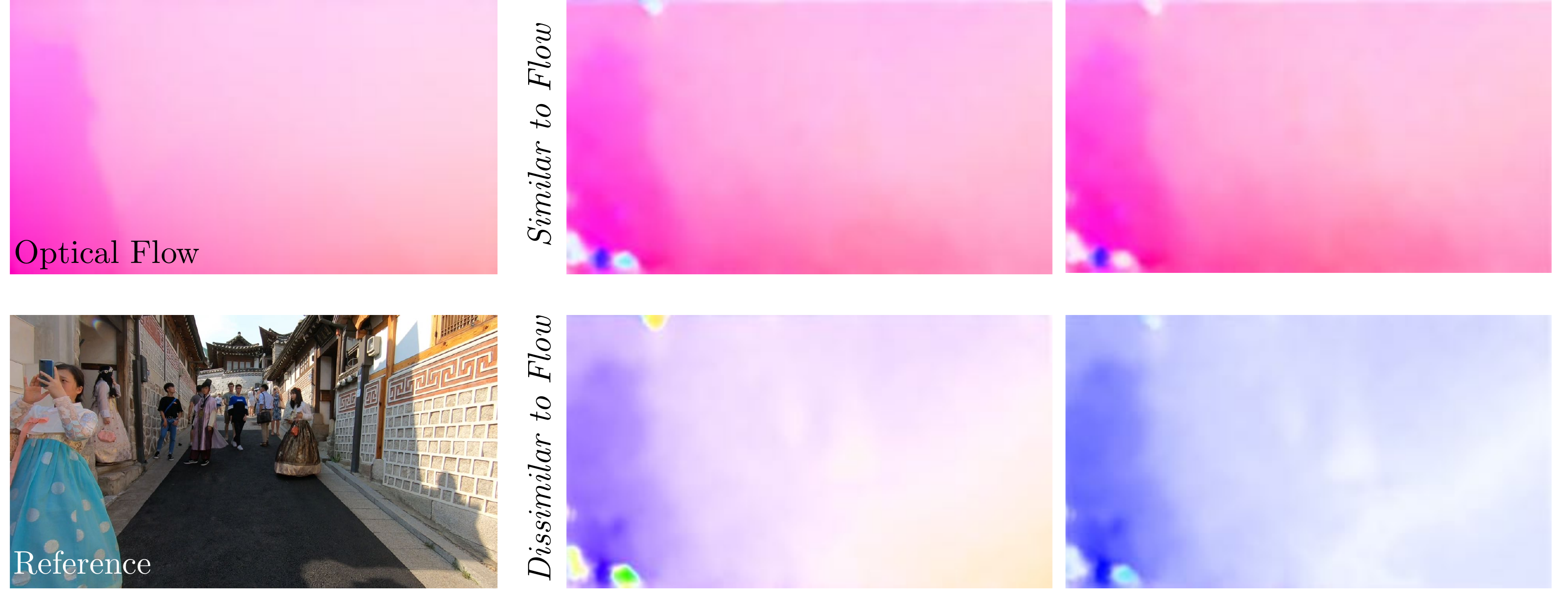}}
		\caption{The learned offsets in the case of $G{=}1, N{=}25$. Although different directions  of offsets are estimated, the learned offsets are all analogous to optical flow in terms of the overall shape.}
		\label{appxfig:offsets}
	\end{center}
\end{figure*}

\begin{figure*}[!h]
	\begin{center}
		\includegraphics[width=0.9\textwidth]{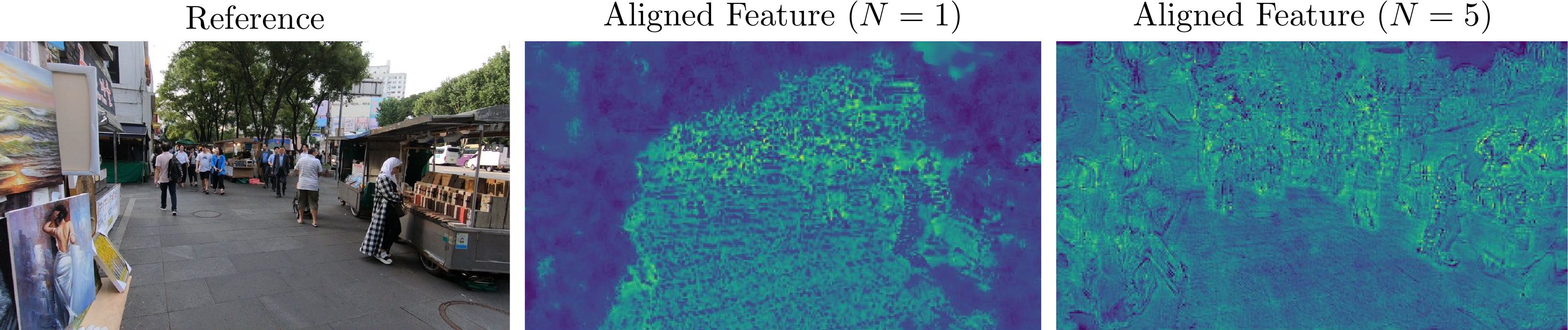}
		\caption{Comparison of the aligned features for $N{=}1$ and $N{=}15$. The aligned feature with multiple offsets are more coherent.}
		\label{appxfig:appx_aligned_fea}
	\end{center}
\end{figure*}

We further visualize the learned offsets when the number of offsets $N$ increases. From Fig.~\ref{appxfig:offsets} we see that when $N$ is increased from $1$ to $25$, the learned offsets are still similar to the optical flow in terms of the overall shape, although they have different estimated directions. These diverse offsets are complementary to each other. As discussed in Sec.~\ref{subsec:flowbased}, these offsets are able to reduce the warping error by aggregating information from different locations, leading to a better aligned feature. Another example is shown in Fig.~\ref{appxfig:appx_aligned_fea}. We see that with only one offset, the learned offsets are unable to locate the correct correspondence (due to a large rotation in the image), resulting in a large area of dark regions. In contrast, when $N{=}5$, the offsets are able to retrieve information from other feasible locations, yielding a more coherent feature.

\begin{figure*}[!h]
	\begin{center}
		\includegraphics[width=\textwidth]{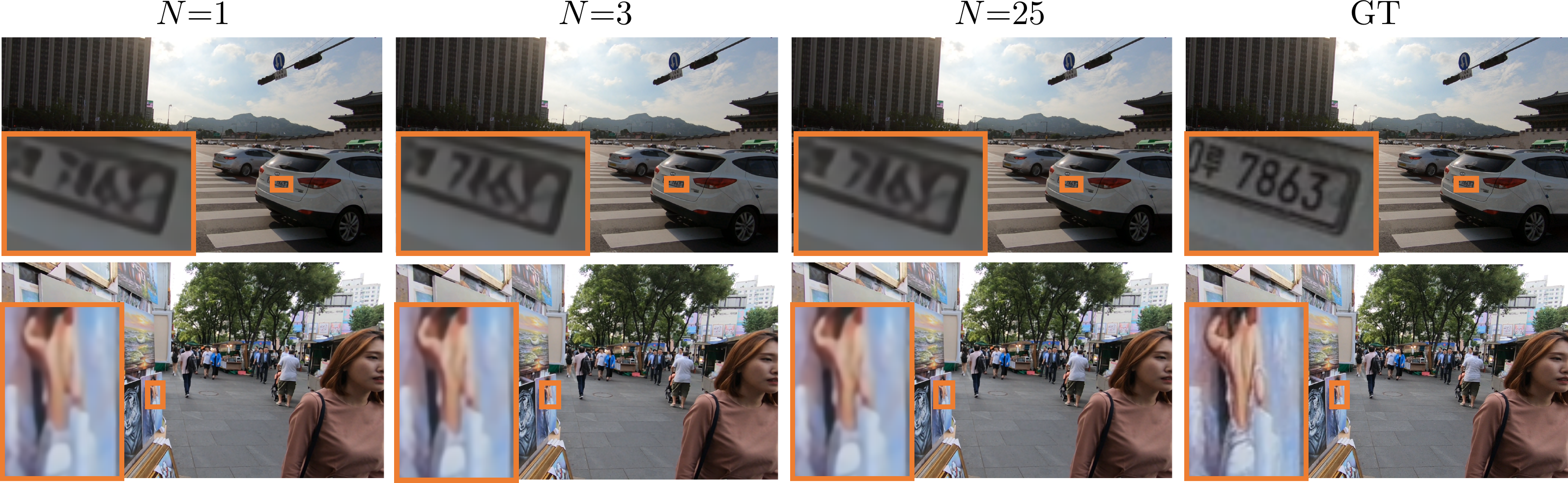}
		\caption{Output HR frames. Quality improvement is observed from $N{=}1$ to $N{=}3$; further improvement is not observed from $N{=}3$ to $N{=}25$.}
		\label{appxfig:N_comp}
	\end{center}
\end{figure*}

The additional information and more coherent features eventually introduce a marked performance gain in video SR. However, it is worth mentioning that indefinitely increasing the offset diversity can reduce the efficiency without significant performance boost. For example, as shown in Fig.~\ref{appxfig:N_comp}, while a significant improvement is observed when traversing from flow-based alignment to deformable alignment ($N{=}3$), further increasing $N$ from $3$ to $25$ does not introduce much visual improvement. The quantitative transition is shown in Fig.~\ref{appxfig:PSNR_heatmap}. We see that when compared to the case when $N{=}1$, using $N{=}5$ offsets leads to an increase of 2.89~dB per 1M parameters. But when $N{=}25$ offsets are used, only 0.42~dB improvement per one million parameters is observed. In practice, one should balance the efficiency and performance by adjusting $N$.

\subsection{Image Warping vs. Feature Warping}
\label{appx:warping}
Another major difference between flow-based alignment and deformable alignment is that flow-based alignment is performed on image level. When aligning images, the spatial warping inevitably introduces information loss, especially high-frequency details, due to the interpolation operation induced by decimal flows (offsets).
On the contrary, this undesirable effect can be alleviated when the alignment is shifted to the feature level.

\begin{figure*}[!h]
	\begin{center}
		\includegraphics[width=\textwidth]{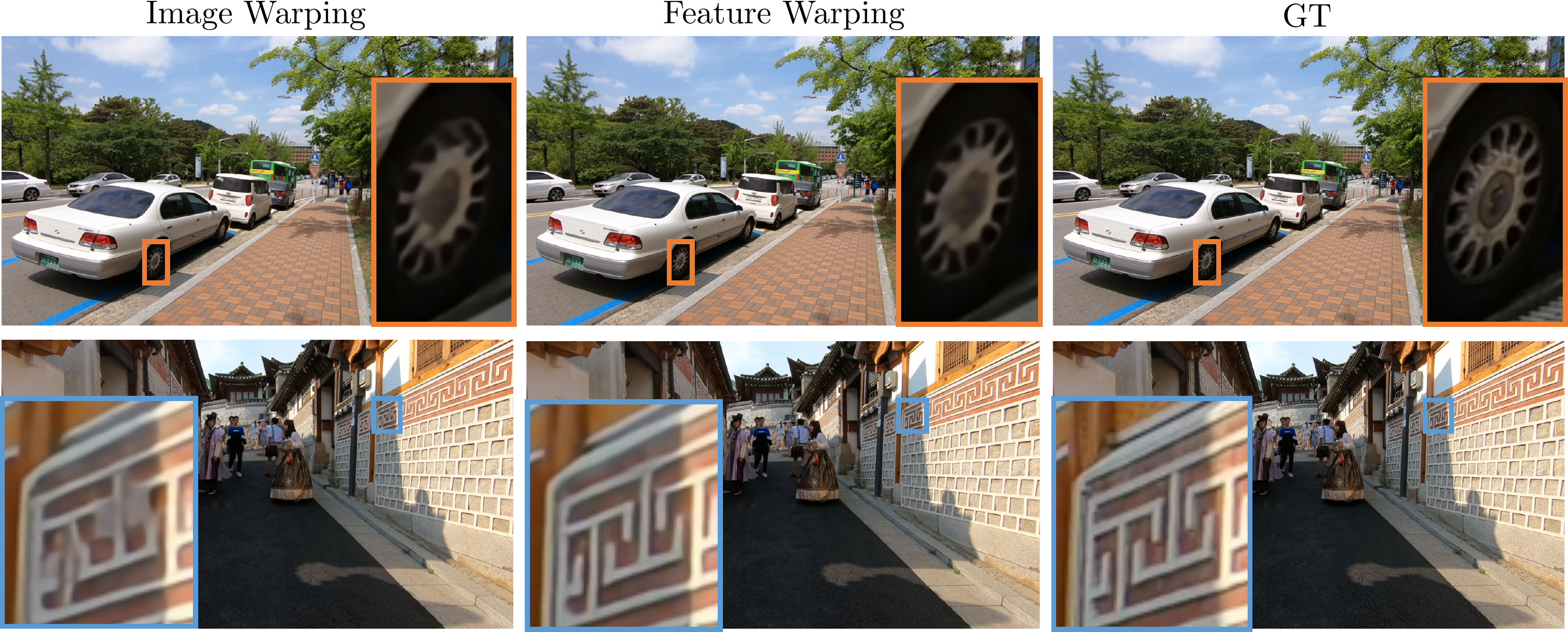}
		\caption{Qualitative comparison. When compared to image warping, feature warping is able to recover more details.}
		\label{appxfig:image_warp}
	\end{center}
\end{figure*}

In this section, we provide additional results to illustrate this another source of superiority of deformable alignment over flow-based alignment. As depicted in Fig.~\ref{appxfig:image_warp}, when compared to image warping, feature warping is able to recover more details. For example, the patterns on the wall are not able to be recovered in image warping. It is surprising that despite the similarity of their architecture, feature warping leads to a improvement of 0.84~dB, as shown in Table~\ref{tab:image_warp}.

\begin{table}[!h]
	\caption{Quantitative comparison between image warping and feature warping on REDS4.}
	\label{tab:image_warp}
	\begin{center}
		\tabcolsep=0.1cm
		\scalebox{1}{
			\begin{tabular}{|r|c|c|c|}
				\hline
				&Image warping&Feature warping&Difference\\\hline
				PSNR (dB)&29.155&29.992&+0.837\\\hline
			\end{tabular}}
	\end{center}
\end{table}

\subsection{Deformable Groups}
\label{appx:groups}
Recall the formulation of deformable alignment:
\begin{equation}
	\hat{F}_{t+i}(\bm{p}) = \sum_{k=1}^{n^2} w(\bm{p}_k) \cdot F_{t+i}(\bm{p}+\bm{p}_k+\Delta\bm{p}_k).
\end{equation}

{\noindent}In practice, one can divide the input channels into $G$ groups, and learn different offsets for different groups. In this case, a total of $G{\times}N$ sets of offsets are learned. Therefore, we see that increasing $G$ is another way to introduce offset diversity. Here, we will discuss in detail the entangled effects of $G$ and $N$.

\begin{figure*}[!h]
	\begin{center}
		\includegraphics[width=0.67\textwidth]{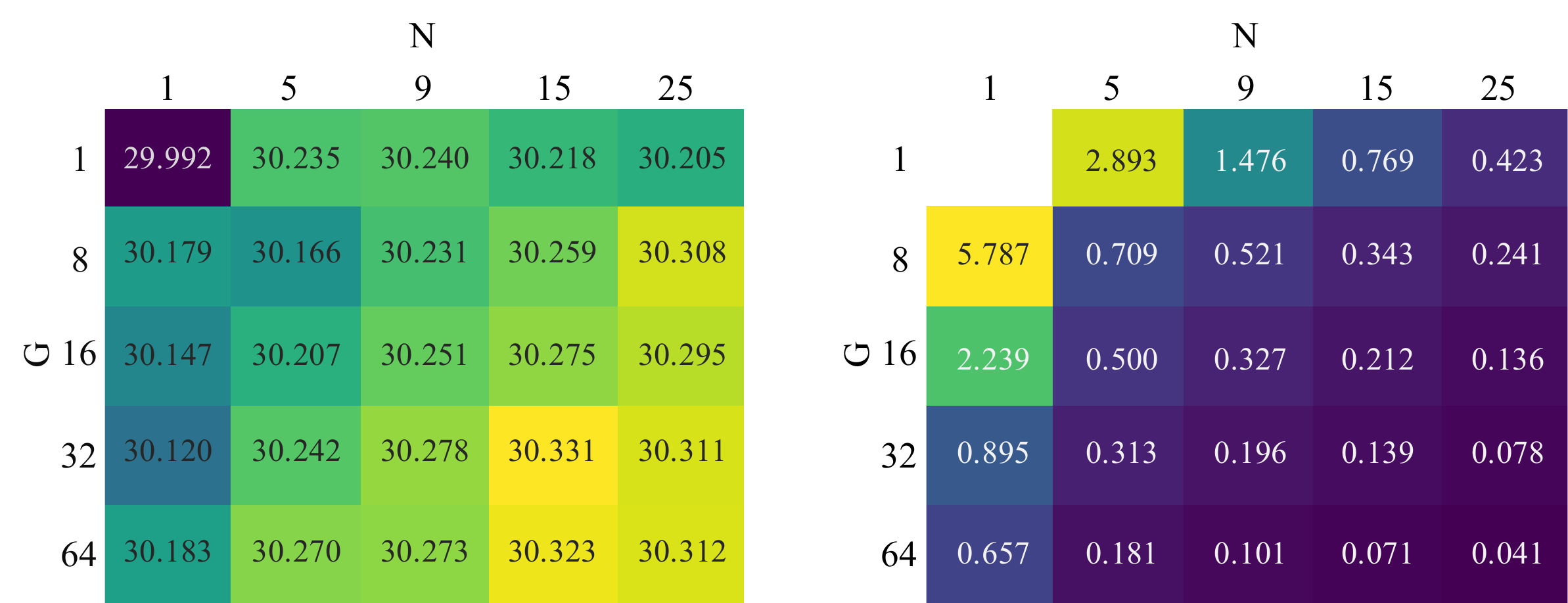}
		\caption{\textit{Left:} the PSNR on REDS4 with different $G$ and $N$. \textit{Right:} PSNR improvement brought by a unit of parameters with different $G$ and $N$ (the numbers are multiplied by 1M for visualization).}
		\label{appxfig:PSNR_heatmap}
	\end{center}
\end{figure*}

The PSNR of models with different $G$ and $N$ are summarized in Fig.~\ref{appxfig:PSNR_heatmap} (\textit{Left}). In general, we see that the PSNR imrpoves with $G$, with up to 0.19 dB. It is consistent with our conclusion for $N$ that the performance positively correlates with the offset diversity.

Moreover, we also observe a tendency of saturation when either $G$ or $N$ become large. In particular, for $G{=}1$, increasing $N$ from 1 to 5 leads to an increase in PSNR by 0.25~dB with only 84K more parameters.
However, when we increase the number of learned offsets up to $64{\times}25{=}1600$ ($G{=}64,N{=}25$), the performance fluctuates at about 30.3~dB. In particular, for $N{=}25$, the performance of $G{=}32$ and $G{=}64$ have nearly identical PSNR, but the latter has 4M more parameters. Figure~\ref{appxfig:PSNR_heatmap} (\textit{Right}) shows the PSNR improvement brought by a unit of parameters with different $G$ and $N$. It is observed that the diversity efficiency drops drastically as $G$ and $N$ increase.

\subsection{Modulation Masks}
\label{appx:masks}
In Modulated deformable convolution (\textit{a.k.a.} DCNv2)~\cite{zhu2019deformable}, an additional mask $\bm{m}_k(\bm{p})$ is learned to further strengthen the capability in manipulating spatial support regions:
\begin{equation}
\label{eq:m_dcn}
    y(\bm{p}) = \sum_{k=1}^{n^2} w(\bm{p}_k) \cdot x\left(\bm{p} + \bm{p}_k + \Delta\bm{p}_k\right)\cdot\bm{m}_k(\bm{p}).
\end{equation}

\begin{figure*}[!h]
	\begin{center}
		\subfloat[$G{=}8, N{=}9$]{\includegraphics[width=0.4\textwidth]{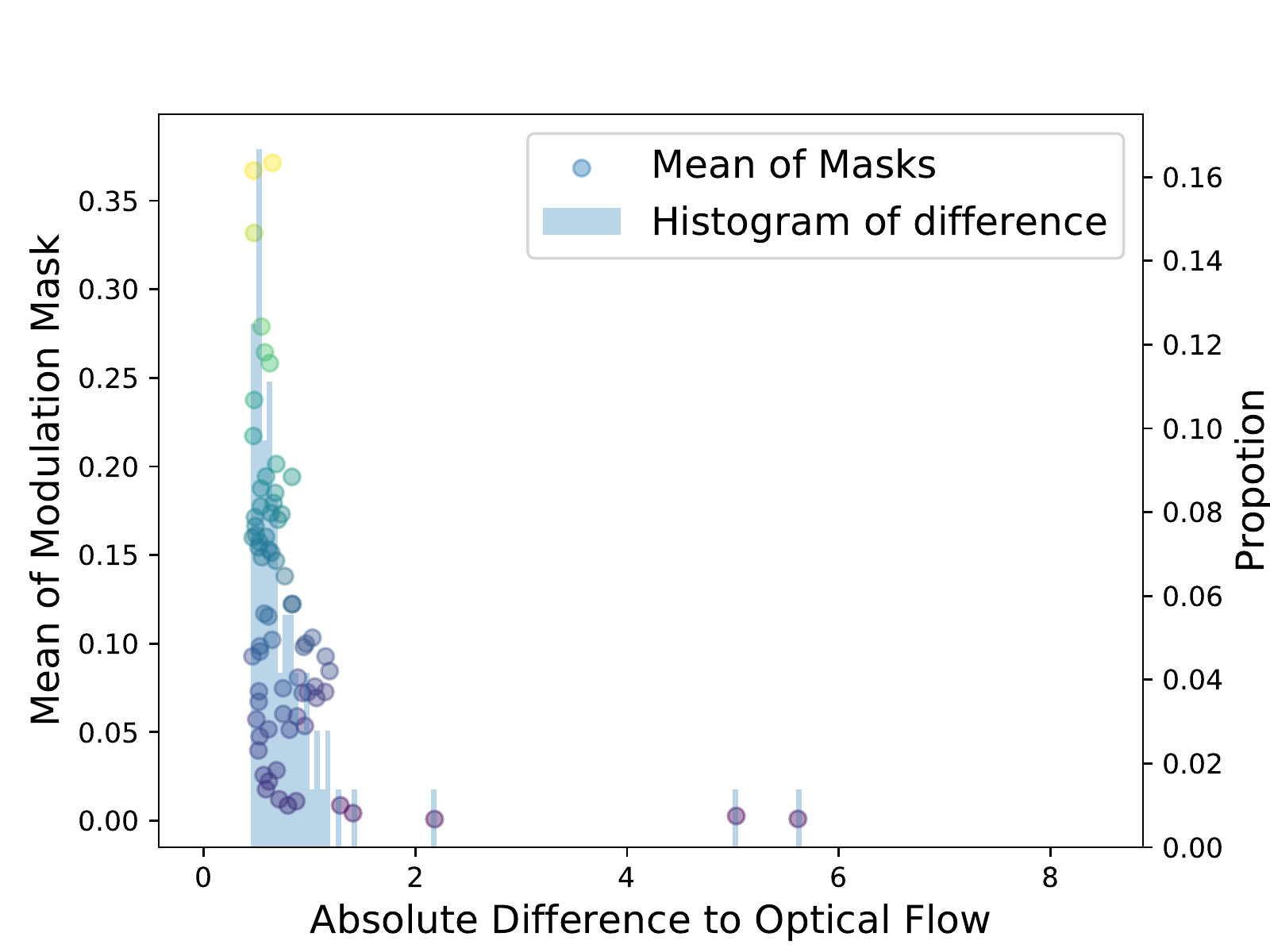}}
		\subfloat[$G{=}64, N{=}25$]{\includegraphics[width=0.4\textwidth]{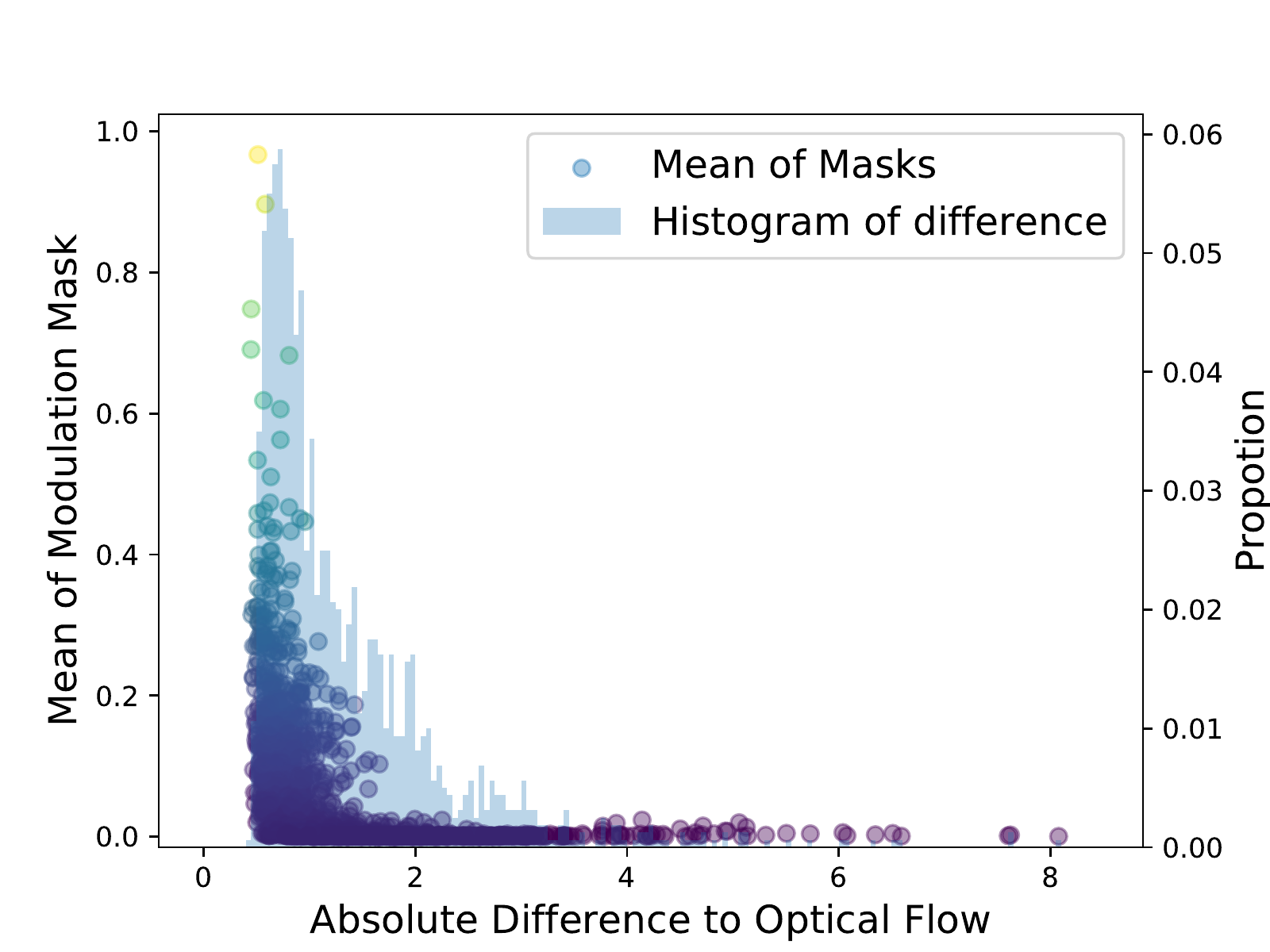}}
		\caption{The average difference against the mean of modulated mask. It is observed from the two plots that (1) the offsets that deviate much from the optical flow have the smallest mask values, and (2) when the number of offsets is large, a significant proportion of the masks has small values ($\approx 0$).}
		\label{appxfig:mask_scatter}
	\end{center}
\end{figure*}

\begin{figure*}[!h]
	\begin{center}
		\includegraphics[width=0.9\textwidth]{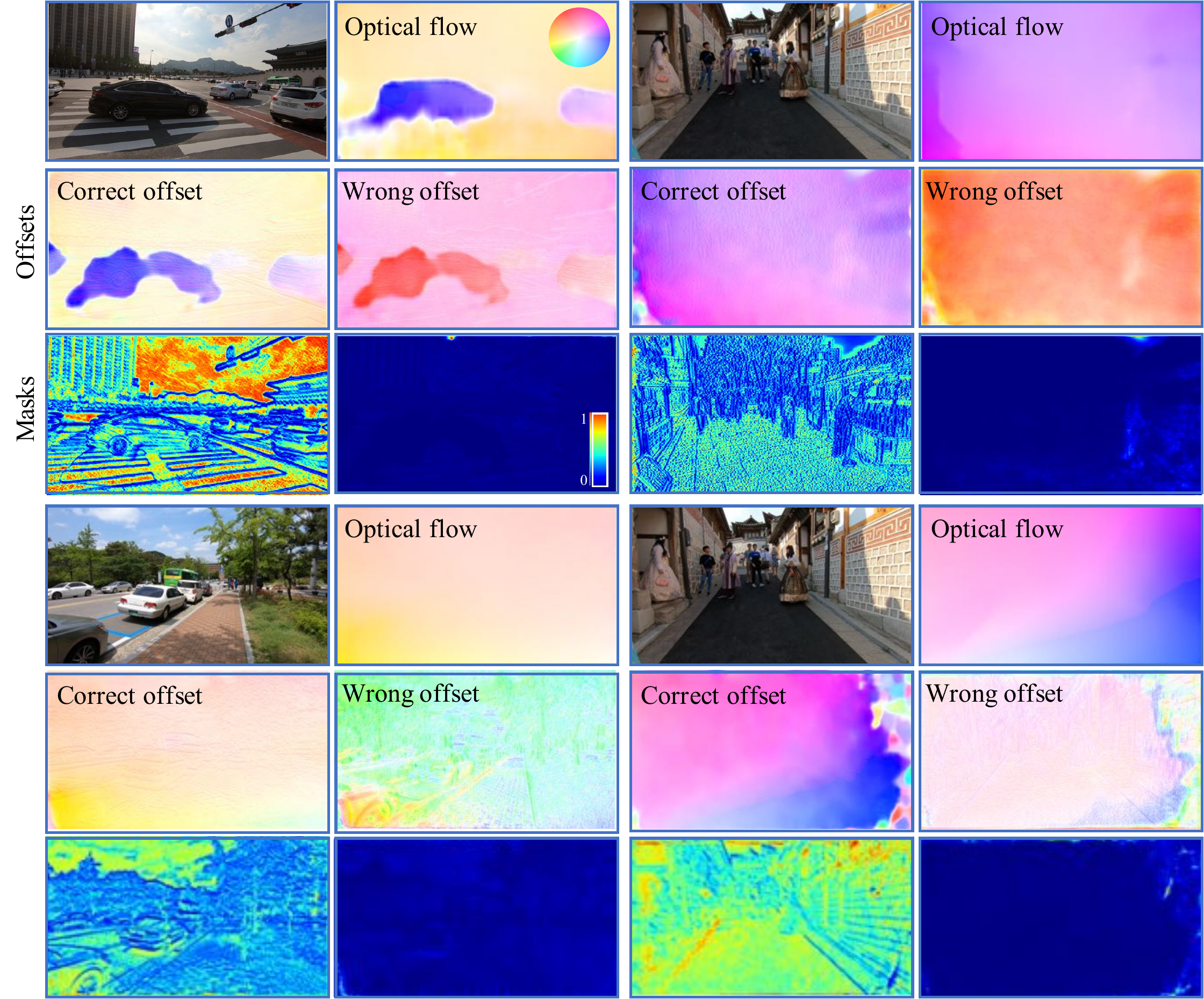}
		\caption{Visualization of offsets and masks. The offsets that are signifiantly different to the optical flow have a relatively small attention in masks.}
		\label{appxfig:offset_mask_comp}
	\end{center}
\end{figure*}

{\noindent}In the presence of modulation masks, the independently warped features are multiplied by the modulation masks before fusion. In this section, we discuss the masks associated to the diverse flow-like offsets.
In Fig.~\ref{appxfig:mask_scatter}, we plot the average difference to optical flow against the mean value of the masks. It is observed that the mean value is inversely proportional to the average difference between the offsets and optical flow. In other words, the offsets that are significantly different to optical flow have relatively small contributions during fusion (see Fig.~\ref{appxfig:offset_mask_comp}).

Furthermore, it is observed that when the number of offsets is large, a significant propotion of the masks has samll values ($\approx 0$). We hypothsize that with a large number of offsets, some offsets become redundant. This agrees with the observation in our main paper that the PSNR tend to saturate when further increasing $G$ or $N$. The modulation mask can be regarded as an attention module that takes the relative importance of the features into account.

\subsection{Offset-Fidelity Loss}
\label{appx:loss}
It is known that the training of deformable alignment is unstable. The unstable training process leads to the overflow of offsets. As a result, the video SR model degenerates to a single image SR model\footnote{As confirmed by the authors of~\cite{wang2019edvr}, the training using the official scheme (\url{https://github.com/xinntao/EDVR}) is unstable and the offsets are prone to overflow. The same result is also observed in our experiments.}~\cite{wang2019edvr}.

To avoid offset overflow, we propose the offset-fidelity loss to constraint the learned offsets so that they cannot deviate much from the optical flow. In addition, as offset diversity has been shown important in video SR, we allow the network to learn the optimal offsets as long as their difference to optical flow does not exceed a certain threshold.

In this work, we conduct experiments on EDVR, an state-of-the-art model using deformable alignment. As discussed in Sec.~\ref{subsec:offset-fidelity}, without the offset-fidelity loss, the overflow of offsets results in a zero feature map after deformable alignment. As a result, EDVR essentially becomes a single image SR model, resulting in a much worse quality. In contrast, with our loss, the network learns interpretable offsets and is able to aggregate the information from neighboring frames, hence achieving significant improvements.

\begin{figure*}[!h]
	\begin{center}
		\subfloat{\includegraphics[height=0.38\textwidth]{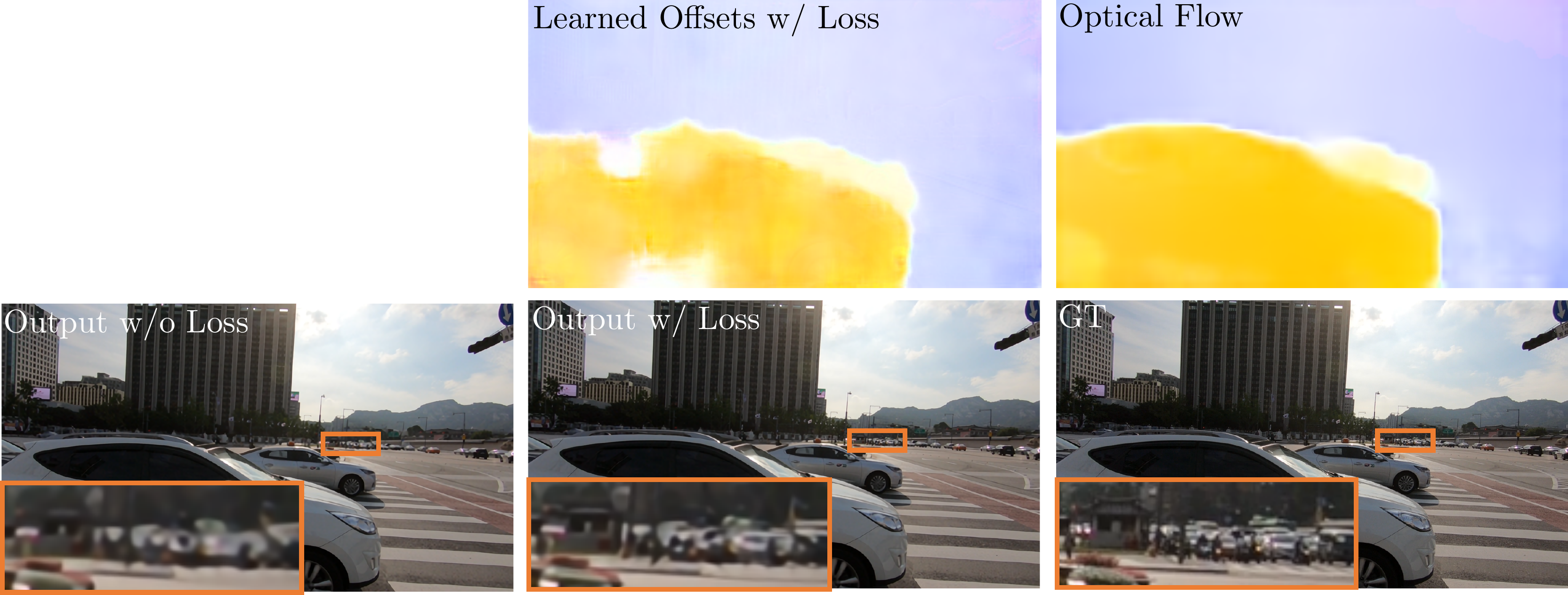}}\\
		\subfloat{\includegraphics[height=0.38\textwidth]{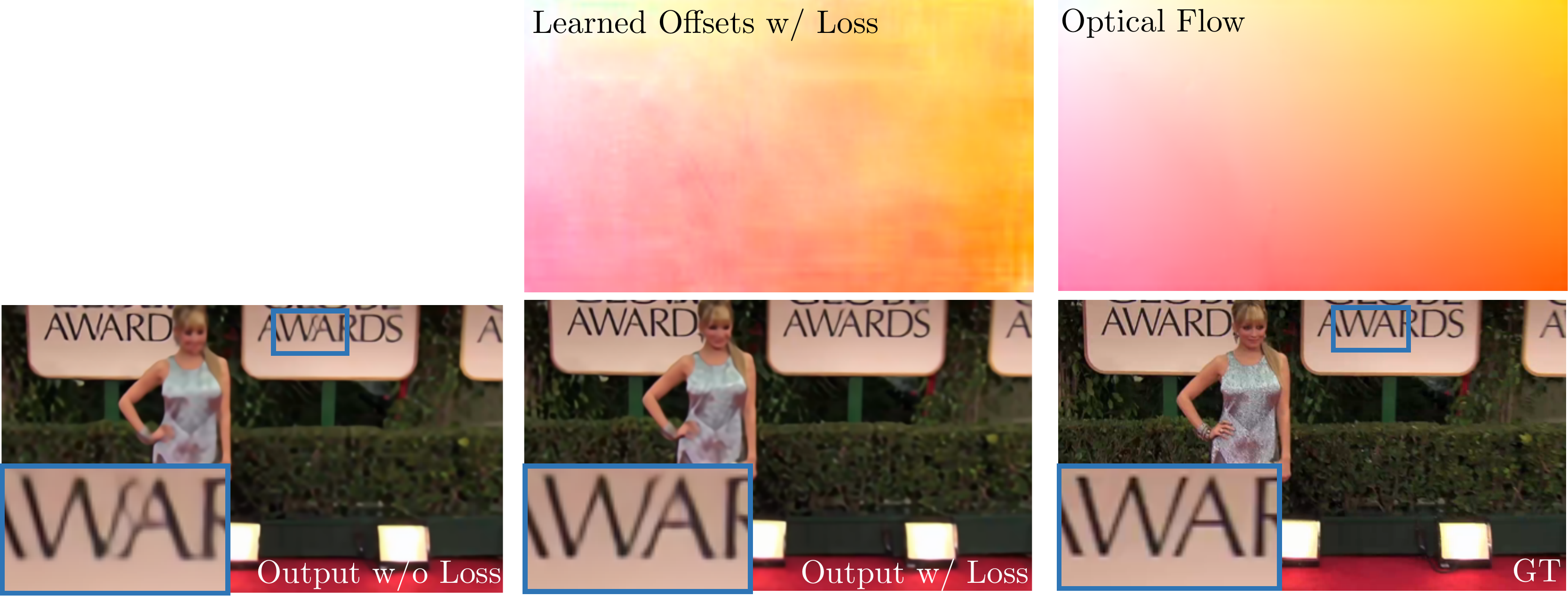}}
		\caption{The results trained with and without offset-fidelity loss.  Without our loss, the overflow of offsets essentially turns EDVR into a single image SR model (The overflowed offsets are not shown). In contrast, our loss successfully stabilizes the training, yielding more interpretable offsets and better output quality}
		\label{appxfig:loss_output}
	\end{center}
\end{figure*}

In this section, we present qualitative results of our offset-fidelity loss. From Fig.~\ref{appxfig:loss_output} we see that with the offset-fidelity loss, EDVR is able to learn interpretable offsets. On the contrary, the offsets trained without the offset-fidelity loss overflows, and the information from neighboring frames cannot be utilized, resulting in blurrier outputs.

\onecolumn
\subsection{Figures}
\clearpage


\twocolumn
\small
\bibliography{short,bib}
\end{document}